%% file: arxiv.tex
\documentclass[10pt,twocolumn,letterpaper]{article}

\usepackage[pagenumbers]{cvpr} 

\usepackage{graphicx}
\usepackage{amsmath}
\usepackage{amssymb}
\usepackage{booktabs}
\usepackage{url}
\usepackage{epsfig}
\usepackage{lipsum}
\usepackage{amsthm}
\usepackage{comment}
\usepackage{multirow}
\usepackage{subcaption}
\usepackage{microtype}
\usepackage{xspace}
\usepackage{setspace}
\usepackage[dvipsnames]{xcolor}
\usepackage[utf8]{inputenc}
\usepackage{enumitem}
\usepackage{makecell}

\usepackage[pagebackref,breaklinks,colorlinks]{hyperref}

\usepackage[capitalize]{cleveref}
\crefname{section}{Sec.}{Secs.}
\Crefname{section}{Section}{Sections}
\Crefname{table}{Table}{Tables}
\crefname{table}{Tab.}{Tabs.}

\def\cvprPaperID{???} 
\def\confName{CVPR}
\def\confYear{2023}

\input{macros.tex}

\begin{document}

\title{\name: High-Fidelity Neural Surface Reconstruction}

\author{
Zhaoshuo Li\textsuperscript{1,2}
\quad Thomas Müller\textsuperscript{1} \quad Alex Evans\textsuperscript{1} \quad Russell H. Taylor\textsuperscript{2} \quad Mathias Unberath\textsuperscript{2} \\ Ming-Yu Liu\textsuperscript{1} \quad Chen-Hsuan Lin\textsuperscript{1}
\vspace{0.2cm} \\
\textsuperscript{1}NVIDIA Research \quad
\textsuperscript{2}Johns Hopkins University \vspace{2pt} \\
{\fontsize{10}{10}\selectfont \url{https://research.nvidia.com/labs/dir/neuralangelo}}
}

\twocolumn[{
    \renewcommand\twocolumn[1][]{#1}
    \maketitle
    \begin{center}
        \input{figures/teaser.tex}
    \end{center}
}]

\begin{abstract}
\input{sections/0-abstract}
\end{abstract}

\input{sections/1-introduction}
\input{sections/2-relatedwork}

\input{sections/3-approach}

\input{sections/4-experiments}

\input{sections/5-conclusion}

\clearpage
\newpage
\input{sections/6-appendix}

{\small
\bibliographystyle{ieee_fullname}
\bibliography{ref}
}

\end{document}

%% file: macros.tex
\newcommand{\name}{{Neuralangelo}\xspace}

\newcommand{\dir}{\mathbf{d}}
\newcommand{\camcenter}{\mathbf{o}}
\newcommand{\rgb}{\mathbf{c}}
\newcommand{\pos}{\mathbf{x}}
\newcommand{\feat}{\gamma}
\newcommand{\tnt}{Tanks and Temples\xspace}
\newcommand{\deriv}[2]{\frac{\partial #1}{\partial #2}}
\newcommand{\sdf}{f}

\newcommand{\lcurv}{\mathcal{L}_\text{curv}}
\newcommand{\leik}{\mathcal{L}_\text{eik}}
\newcommand{\lrgb}{\mathcal{L}_\text{RGB}}

\newcommand{\SFM}{S\textit{f}M\xspace}

\newcommand{\agl}{AG+P\xspace}
\newcommand{\ag}{AG\xspace}
\newcommand{\fd}{NG\xspace}
\newcommand{\fdl}{NG+P\xspace}

\newcommand{\best}{\color{OliveGreen} \bf}
\newcommand{\sbest}{\color{NavyBlue} \bf}

\usepackage{booktabs}
\newcommand{\ra}[1]{\renewcommand{\arraystretch}{#1}}
\usepackage[accsupp]{axessibility}
\usepackage{nimbusmononarrow}

%% file: figures/teaser.tex
    \vspace{-12pt}
    \centering
    \includegraphics[width=\linewidth]{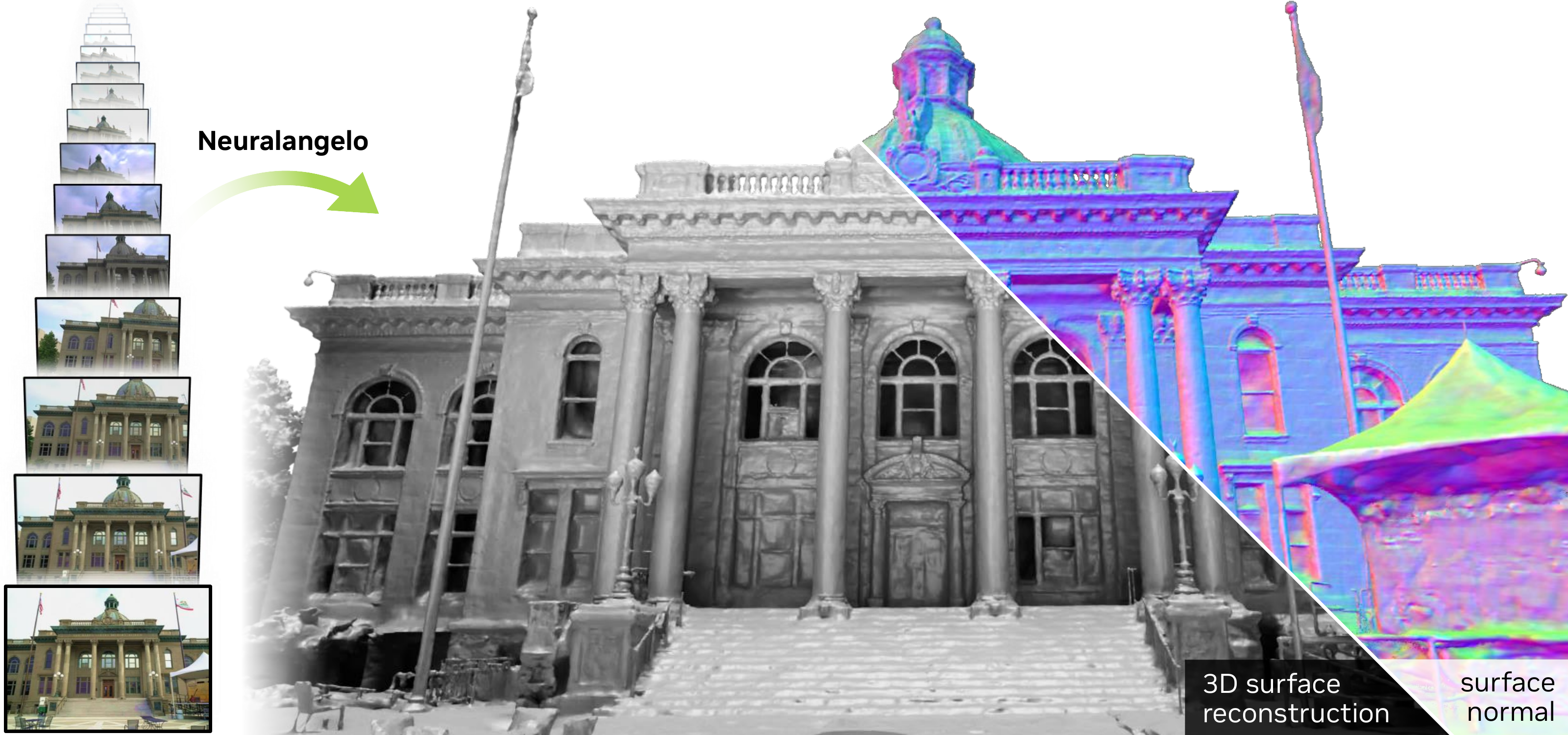}
    \captionof{figure}{
        We present \textbf{\name}, a framework for high-fidelity 3D surface reconstruction from RGB images using neural volume rendering, even without auxiliary data such as segmentation or depth.
        Shown in the figure is an extracted 3D mesh of a courthouse.
    }
    \label{fig:teaser}
    \vspace{8pt}

%% file: sections/0-abstract.tex
\vspace{-4pt}

Neural surface reconstruction has been shown to be powerful for recovering dense 3D surfaces via image-based neural rendering.
However, current methods struggle to recover detailed structures of real-world scenes.
To address the issue, we present \textbf{\name}, which combines the representation power of multi-resolution 3D hash grids with neural surface rendering.
Two key ingredients enable our approach: (1) numerical gradients for computing higher-order derivatives as a smoothing operation and (2) coarse-to-fine optimization on the hash grids controlling different levels of details.
Even without auxiliary inputs such as depth, \name can effectively recover dense 3D surface structures from multi-view images with fidelity significantly surpassing previous methods, enabling detailed large-scale scene reconstruction from RGB video captures.

\vspace{-8pt}

%% file: sections/1-introduction.tex
\section{Introduction}
\label{sec:intro}

3D surface reconstruction aims to recover dense geometric scene structures from multiple images observed at different viewpoints~\cite{hartley2003multiple}.
The recovered surfaces provide structural information useful for many downstream applications, such as 3D asset generation for augmented/virtual/mixed reality or environment mapping for autonomous navigation of robotics.
Photogrammetric surface reconstruction using a monocular RGB camera is of particular interest, as it equips users with the capability of casually creating digital twins of the real world using ubiquitous mobile devices.

Classically, multi-view stereo algorithms~\cite{seitz1999photorealistic, kutulakos2000theory,furukawa2009accurate,tola2012efficient} had been the method of choice for sparse 3D reconstruction.
An inherent drawback of these algorithms, however, is their inability to handle ambiguous observations, \eg regions with large areas of homogeneous colors, repetitive texture patterns, or strong color variations. 
This would result in inaccurate reconstructions with noisy or missing surfaces. 
Recently, neural surface reconstruction methods~\cite{yariv2020multiview,yariv2021volume,wang2021neus} have shown great potential in addressing these limitations. 
This new class of methods uses coordinate-based multi-layer perceptrons (MLPs) to represent the scene as an implicit function, such as occupancy fields~\cite{oechsle2021unisurf} or signed distance functions (SDF)~\cite{yariv2020multiview,yariv2021volume,wang2021neus}.
Leveraging the inherent continuity of MLPs and neural volume rendering~\cite{mildenhall2020nerf}, these techniques allow the optimized surfaces to meaningfully interpolate between spatial locations, resulting in smooth and complete surface representations.

Despite the superiority of neural surface reconstruction methods over classical approaches, the recovered fidelity of current methods does not scale well with the capacity of MLPs.
Recently, M\"{u}ller~\etal~\cite{muller2022instant} proposed a new scalable representation, referred to as Instant NGP (Neural Graphics Primitives). 
Instant NGP introduces a hybrid 3D grid structure with a multi-resolution hash encoding and a lightweight MLP that is more expressive with a memory footprint log-linear to the resolution.
The proposed hybrid representation greatly increases the representation power of neural fields and has achieved great success at representing very fine-grained details for a wide variety of tasks, such as object shape representation and novel view synthesis problems.

In this paper, we propose \textbf{\name} for high-fidelity surface reconstruction (Fig.~\ref{fig:teaser}). 
\name adopts Instant NGP as a neural SDF representation of the underlying 3D scene, optimized from multi-view image observations via neural surface rendering~\cite{wang2021neus}.
We present two findings central to fully unlocking the potentials of multi-resolution hash encodings.
First, using \emph{numerical} gradients to compute higher-order derivatives, such as surface normals for the eikonal regularization~\cite{gropp2020implicit,yariv2020multiview,jiang2020sdfdiff,lin2020sdf}, is critical to stabilizing the optimization.
Second, a \emph{progressive} optimization schedule plays an important role in recovering the structures at different levels of details. 
We combine these two key ingredients and, via extensive experiments on standard benchmarks and real-world scenes, demonstrate significant improvements over image-based neural surface reconstruction methods in \emph{both} reconstruction accuracy and view synthesis quality.

In summary, we present the following contributions:

\begin{itemize}[leftmargin=18pt]
    \setlength\itemsep{0pt}
    \item We present the \name framework to naturally incorporate the representation power of multi-resolution hash encoding~\cite{muller2022instant} into neural SDF representations.
    \item We present two simple techniques to improve the quality of hash-encoded surface reconstruction: higher-order derivatives with numerical gradients and coarse-to-fine optimization with a progressive level of details.
    \item We empirically demonstrate the effectiveness of \name on various datasets, showing significant improvements over previous methods.
\end{itemize}

%% file: sections/2-relatedwork.tex
\section{Related work}
\label{sec:related_work}

\vspace{4pt}
\noindent\textbf{Multi-view surface reconstruction.}
Early image-based photogrammetry techniques use a volumetric occupancy grid to represent the scene~\cite{szeliski1993rapid, laurentini1994visual, de1999poxels, seitz1999photorealistic, kutulakos2000theory}.
Each voxel is visited and marked occupied if strict color constancy between the corresponding projected image pixels is satisfied.
The photometric consistency assumption typically fails due to auto-exposure or non-Lambertian materials, which are ubiquitous in the real world.
Relaxing such color constancy constraints across views is important for realistic 3D reconstruction.

Follow-up methods typically start with 3D point clouds from multi-view stereo techniques~\cite{furukawa2009accurate,tola2012efficient,galliani2015massively,schonberger2016pixelwise} and then perform dense surface reconstruction~\cite{kazhdan2006poisson,kazhdan2013screened}.
Reliance on the quality of the generated point clouds often leads to missing or noisy surfaces.
Recent learning-based approaches augment the point cloud generation process with learned image features and cost volume construction~\cite{huang2018deepmvs, yao2018mvsnet, chen2019point}.
However, these approaches are inherently limited by the resolution of the cost volume and fail to recover geometric details.

\vspace{4pt}
\noindent\textbf{Neural Radiance Fields (NeRF).}
NeRF~\cite{mildenhall2020nerf} achieves remarkable photorealistic view synthesis with view-dependent effects.
NeRF encodes 3D scenes with an MLP mapping 3D spatial locations to color and volume density.
These predictions are composited into pixel colors using neural volume rendering.
A problem of NeRF and its variants~\cite{zhang2020nerf++,barron2021mip,yu2021plenoxels,sun2022direct}, however, is the question of how an isosurface of the volume density could be defined to represent the underlying 3D geometry.
Current practice often relies on heuristic thresholding on the density values; due to insufficient constraints on the level sets, however, such surfaces are often noisy and may not model the scene structures accurately~\cite{yariv2021volume,wang2021neus}.
Therefore, more direct modeling of surfaces is preferred for photogrammetric surface reconstruction problems.

\vspace{4pt}
\noindent\textbf{Neural surface reconstruction.} 
For scene representations with better-defined 3D surfaces, implicit functions such as occupancy grids~\cite{DVR,oechsle2021unisurf} or SDFs~\cite{yariv2020multiview} are preferred over simple volume density fields. 
To integrate with neural volume rendering~\cite{mildenhall2020nerf}, different techniques~\cite{yariv2021volume,wang2021neus} have been proposed to reparametrize the underlying representations back to volume density.
These designs of neural implicit functions enable more accurate surface prediction with view synthesis capabilities of unsacrificed quality~\cite{yariv2020multiview}.

Follow-up works extend the above approaches to real-time at the cost of surface fidelity~\cite{li2022vox,wang2022neus2}, while others~\cite{darmon2022improving,fu2022geo,yu2022monosdf} use auxiliary information to enhance the reconstruction results.
Notably, NeuralWarp~\cite{darmon2022improving} uses patch warping given co-visibility information from structure-from-motion (\SFM) to guide surface optimization, but the patch-wise planar assumption fails to capture highly-varying surfaces~\cite{darmon2022improving}.
Other methods~\cite{fu2022geo,zhang2022critical} utilize sparse point clouds from \SFM to supervise the SDF, but their performances are upper-bounded by the quality of the point clouds, as with classical approaches~\cite{zhang2022critical}.
The use of depth and segmentation as auxiliary data has also been explored with unconstrained image collections~\cite{sun2022neural} or using scene representations with hash encodings~\cite{yu2022monosdf, zhao2022human}.
In contrast, our work Neuralangelo builds upon hash encodings~\cite{muller2022instant} to recover surfaces but \emph{without} the need for auxiliary inputs used in prior work~\cite{fu2022geo,zhang2022critical,sun2022neural,yu2022monosdf,darmon2022improving}.
Concurrent work~\cite{wanghf} also proposes coarse-to-fine optimization for improved surface details, where a displacement network corrects the shape predicted by a coarse network.
In contrast, we use hierarchical hash grids and control the level of details based on our analysis of higher-order derivatives.

%% file: sections/3-approach.tex
\section{Approach}

\name reconstructs dense structures of the scene from multi-view images.
\name samples 3D locations along camera view directions and uses a multi-resolution hash encoding to encode the positions.
The encoded features are input to an SDF MLP and a color MLP to composite images using SDF-based volume rendering.

\subsection{Preliminaries}
\noindent\textbf{Neural volume rendering.}
NeRF~\cite{mildenhall2020nerf} represents a 3D scene as volume density and color fields.
Given a posed camera and a ray direction, the volume rendering scheme integrates the color radiance of sampled points along the ray.
The $i$-th sampled 3D position $\pos_i$ is at a distance $t_i$ from the camera center.
The volume density $\sigma_i$ and color $\rgb_i$ of each sampled point are predicted using a coordinate MLP.
The rendered color of a given pixel is approximated as the Riemann sum:
\begin{equation}
    \hat{\rgb}(\camcenter, \dir) = \sum_{i=1}^N w_i \rgb_i, \;\;\; \text{where} \,\, w_i = T_i \alpha_i.
    \label{eqn:volume_rendering}
\end{equation}
Here, $\alpha_i = 1 - \mathrm{exp}(-\sigma_i \delta_i)$ is the opacity of the $i$-th ray segment, $\delta_i = t_{i+1} - t_i$ is the distance between adjacent samples, and $T_i = \Pi_{j=1}^{i-1} (1-\alpha_j)$ is the accumulated transmittance, indicating the fraction of light that reaches the camera.
To supervise the network, a color loss is used between input images $\rgb$ and rendered images $\hat{\rgb}$:
\begin{equation}
    \lrgb = \| \hat{\rgb} - \rgb \|_1 \, .
    \label{eqn:color_loss}
\end{equation}
However, surfaces are not clearly defined using such density formulation.
Extracting surfaces from density-based representation often leads to noisy and unrealistic results~\cite{yariv2021volume,wang2021neus}.

\vspace{4pt}
\noindent\textbf{Volume rendering of SDF.}
One of the most common surface representations is SDF.
The surface $\mathcal{S}$ of an SDF can be implicitly represented by its zero-level set, \ie, $\mathcal{S} = \{\pos \in \mathbb{R}^3 | f(\pos) = 0\}$, where $\sdf(\pos)$ is the SDF value.
In the context of neural SDFs, Wang~\etal~\cite{wang2021neus} proposed to convert volume density predictions in NeRF to SDF representations with a logistic function to allow optimization with neural volume rendering.
Given a 3D point $\pos_i$ and SDF value $\sdf(\pos_i)$, the corresponding opacity value $\alpha_i$ used in Eq.~\ref{eqn:volume_rendering} is computed as
\begin{equation}
    \alpha_i = \mathrm{max} \left( \frac{\Phi_s(\sdf(\pos_i)) - \Phi_s(\sdf(\pos_{i+1}))}{\Phi_s(\sdf(\pos_i))}, 0 \right),
\end{equation}
where $\Phi_s$ is the sigmoid function.
In this work, we use the same SDF-based volume rendering formulation~\cite{wang2021neus}.

\vspace{4pt}
\noindent\textbf{Multi-resolution hash encoding.}
Recently, multi-resolution hash encoding proposed by M\"{u}ller \textit{et al.}~\cite{muller2022instant} has shown great scalability for neural scene representations, generating fine-grained details for tasks such as novel view synthesis.
In \name, we adopt the representation power of hash encoding to recover high-fidelity surfaces. 

The hash encoding uses multi-resolution grids, with each grid cell corner mapped to a hash entry. Each hash entry stores the encoding feature. 
Let $\{V_1, ..., V_L\}$ be the set of different spatial grid resolutions.
Given an input position $\pos_i$, we map it to the corresponding position at each grid resolution $V_l$ as $\pos_{i,l} = \pos_i \cdot V_l$.
The feature vector $\feat_l(\pos_{i,l}) \in \mathbb{R}^{c}$ given resolution $V_l$ is obtained via trilinear interpolation of hash entries at the grid cell corners.
The encoding features across all spatial resolutions are concatenated together, forming a $\feat(\pos_{i}) \in \mathbb{R}^{cL}$ feature vector:
\begin{equation}
    \feat(\pos_{i}) = \big( \feat_1(\pos_{i,1}),...,\feat_L(\pos_{i,L}) \big).
\end{equation}
The encoded features are then passed to a shallow MLP. 

One alternative to hash encoding is sparse voxel structures \cite{yu2021plenoxels,takikawa2021neural,sun2022direct, wu2022voxurf}, where each grid corner is uniquely defined without collision.
However, volumetric feature grids require hierarchical spatial decomposition (\eg octrees) to make the parameter count tractable; otherwise, the memory would grow cubically with spatial resolution. 
Given such hierarchy, finer voxel resolutions by design cannot recover surfaces that are misrepresented by the coarser resolutions~\cite{takikawa2021neural}.
Hash encoding instead assumes no spatial hierarchy and resolves collision automatically based on gradient averaging~\cite{muller2022instant}.

\subsection{Numerical Gradient Computation}
We show in this section that the analytical gradient \wrt position of hash encoding suffers from localities.
Therefore, optimization updates only propagate to local hash grids, lacking non-local smoothness.
We propose a simple fix to such a locality problem by using numerical gradients.
An overview is shown in Fig.~\ref{fig:ag_ng}.

A special property of SDF is its differentiability with a gradient of the unit norm.
The gradient of SDF satisfies the eikonal equation $\|\nabla \sdf(\pos)\|_2 = 1$ (almost everywhere).
To enforce the optimized neural representation to be a valid SDF, the eikonal loss~\cite{gropp2020implicit} is typically imposed on the SDF predictions:
\begin{equation}
    \leik = \frac{1}{N} \sum_{i=1}^N ( \| \nabla \sdf(\pos_i) \|_2 - 1)^2 ,
    \label{eqn:eikonal_loss}
\end{equation}
where $N$ is the total number of sampled points.
To allow for end-to-end optimization, a double backward operation on the SDF prediction $\sdf(\pos)$ is required.

\input{figures/ag_ng.tex}

The \textit{de facto} method for computing surface normals of SDFs $\nabla \sdf(\pos)$ is to use analytical gradients~\cite{wang2021neus,yariv2020multiview,yariv2021volume,yang2021geometry}.
Analytical gradients of hash encoding \wrt position, however, are \emph{not} continuous across space under trilinear interpolation.
To find the sampling location in a voxel grid, each 3D point $\pos_i$ would first be scaled by the grid resolution $V_l$, written as $\pos_{i,l} = \pos_i \cdot V_l$.
Let the coefficient for (tri-)linear interpolation be $\beta = \pos_{i,l} - \lfloor \pos_{i,l} \rfloor$.
The resulting feature vectors are
\begin{equation}
    \feat_l(\pos_{i,l}) = \feat_l(\lfloor \pos_{i,l} \rfloor)  \cdot (1 - \beta) + \feat_l(\lceil \pos_{i,l} \rceil) \cdot \beta ,
\end{equation}
where the rounded position $\lfloor \pos_{i,l} \rfloor, \lceil \pos_{i,l} \rceil$ correspond to the local grid cell corners. 
We note that rounding operations $\lfloor \cdot \rfloor$ and $\lceil \cdot \rceil$ are non-differentiable. 
As a result, the derivative of hash encoding \wrt the position can be obtained as
\begin{align}
    \deriv{\feat_l(\pos_{i,l})}{\pos_i} &=  \feat_l(\lfloor\pos_{i,l}\rfloor) \cdot (- \deriv{\beta}{\pos_{i}}) + \feat_l(\lceil\pos_{i,l} \rceil) \cdot \deriv{\beta}{\pos_{i}} \nonumber \\
    &= \feat_l(\lfloor\pos_{i,l}\rfloor) \cdot (- V_l) + \feat_l(\lceil\pos_{i,l} \rceil) \cdot V_l \;.
\end{align}

The derivative of hash encoding is local, \ie,
when $\pos_i$ moves across grid cell borders, the corresponding hash entries will be different. 
Therefore, the eikonal loss defined in Eq.~\ref{eqn:eikonal_loss} only back-propagates to the locally sampled hash entries, \ie $\feat_l(\lfloor\pos_{i,l}\rfloor)$ and $\feat_l(\lceil\pos_{i,l} \rceil)$. 
When continuous surfaces (\eg a flat wall) span multiple grid cells, these grid cells should produce coherent surface normals without sudden transitions. 
To ensure consistency in surface representation, joint optimization of these grid cells is desirable.
However, the analytical gradient is limited to local grid cells, unless all corresponding grid cells happen to be sampled and optimized simultaneously.
Such sampling is not always guaranteed.

To overcome the locality of the analytical gradient of hash encoding, we propose to compute the surface normals using \textit{numerical} gradients.
If the step size of the numerical gradient is smaller than the grid size of hash encoding, the numerical gradient would be equivalent to the analytical gradient; otherwise, hash entries of multiple grid cells would participate in the surface normal computation.
Back-propagating through the surface normals thus allows hash entries of multiple grids to receive optimization updates simultaneously. 
Intuitively, numerical gradients with carefully chosen step sizes can be interpreted as a smoothing operation on the analytical gradient expression.
An alternative of normal supervision is a teacher-student curriculum~\cite{verbin2022ref,zhang2021nerfactor}, where
the predicted noisy normals are driven towards MLP outputs to exploit the smoothness of MLPs.
However, analytical gradients from such teacher-student losses
still only back-propagate to local grid cells for hash encoding.
In contrast, numerical gradients solve the locality issue without the need of additional networks.

To compute the surface normals using the numerical gradient, additional SDF samples are needed.
Given a sampled point $\pos_i = (x_i, y_i, z_i)$, we additionally sample two points along each axis of the canonical coordinate around $x_i$ within a vicinity of a step size of $\epsilon$.
For example, the $x$-component of the surface normal can be found as
\begin{equation}
    \nabla_x \sdf(\pos_i) = \frac{\sdf \left( \feat(\pos_i + \boldsymbol{\epsilon}_x) \right) - \sdf \left( \feat(\pos_i - \boldsymbol{\epsilon}_x) \right)}{2 \epsilon} \;,
    \label{eqn:central_diff_normal}
\end{equation}
where $\boldsymbol{\epsilon}_x=[\epsilon, 0, 0]$.
In total, six additional SDF samples are required for numerical surface normal computation.

\input{figures/dtu.tex}

\subsection{Progressive Levels of Details}
Coarse-to-fine optimization can better shape the loss landscape to avoid falling into false local minima. 
Such a strategy has found many applications in computer vision, such as image-based registration~\cite{lucas1981iterative,lin2021barf,park2021nerfies}. 
\name also adopts a coarse-to-fine optimization scheme to reconstruct the surfaces with progressive levels of details.  
Using numerical gradients for the higher-order derivatives naturally enables \name to perform coarse-to-fine optimization from two perspectives.

\vspace{4pt}
\noindent\textbf{Step size $\epsilon$.} 
As previously discussed, numerical gradients can be interpreted as a smoothing operation where the step size $\epsilon$ controls the resolution and the amount of recovered details.
Imposing $\leik$ with a larger $\epsilon$ for numerical surface normal computation ensures the surface normal is consistent at a larger scale, thus producing consistent and continuous surfaces.
On the other hand, imposing $\leik$ with a smaller $\epsilon$ affects a smaller region and avoids smoothing details.
In practice, we initialize the step size $\epsilon$ to the coarsest hash grid size and exponentially decrease it matching different hash grid sizes throughout the optimization process.

\vspace{4pt}
\noindent\textbf{Hash grid resolution $V$.} 
If all hash grids are activated from the start of the optimization, to capture geometric details, fine hash grids must first ``unlearn'' from the coarse optimization with large step size $\epsilon$ and ``relearn'' with a smaller $\epsilon$. 
If such a process is unsuccessful due to converged optimization, geometric details would be lost.
Therefore, we only enable an initial set of coarse hash grids and progressively activate finer hash grids throughout optimization when $\epsilon$ decreases to their spatial size.
The relearning process can thus be avoided to better capture the details. 
In practice, we also apply weight decay over all parameters to avoid single-resolution features dominating the final results.

\subsection{Optimization}
To further encourage the smoothness of the reconstructed surfaces, we impose a prior by regularizing the mean curvature of SDF. 
The mean curvature is computed from discrete Laplacian similar to the surface normal computation, otherwise, the second-order analytical gradients of hash encoding are zero everywhere when using trilinear interpolation.
The curvature loss $\lcurv$ is defined as:
\begin{equation}
    \lcurv = \frac{1}{N} \sum_{i=1}^N \left| \nabla^2 \sdf(\pos_i) \right| \,.
\end{equation}
We note that the samples used for the surface normal computation in Eq.~\ref{eqn:central_diff_normal} are sufficient for curvature computation. 

The total loss is defined as the weighted sum of losses:
\begin{equation}
    \mathcal{L} = \lrgb + w_\text{eik} \leik + w_\text{curv} \lcurv \, .
    \label{eqn:total_loss}
\end{equation}
All network parameters, including MLPs and hash encoding, are trained jointly end-to-end.

%% file: figures/ag_ng.tex
\begin{figure}[bt]
    \centering
    \includegraphics[width=\linewidth]{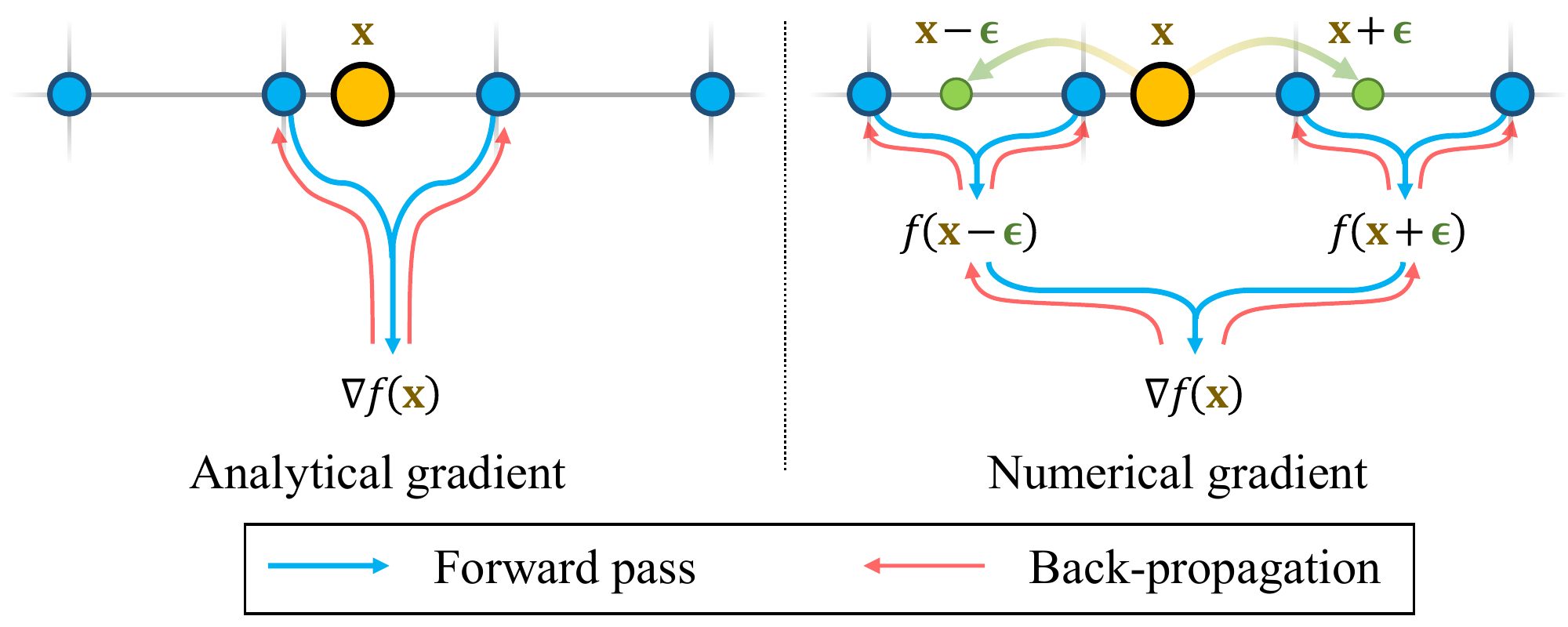}
    \caption{Using \textbf{numerical gradients} for higher-order derivatives distributes the back-propagation updates beyond the local hash grid cell, thus becoming a smoothed version of \textbf{analytical gradients}. 
    }
    \label{fig:ag_ng}
\end{figure}

%% file: figures/dtu.tex
\begin{figure*}[bt]
    \centering
    \includegraphics[width=\linewidth]{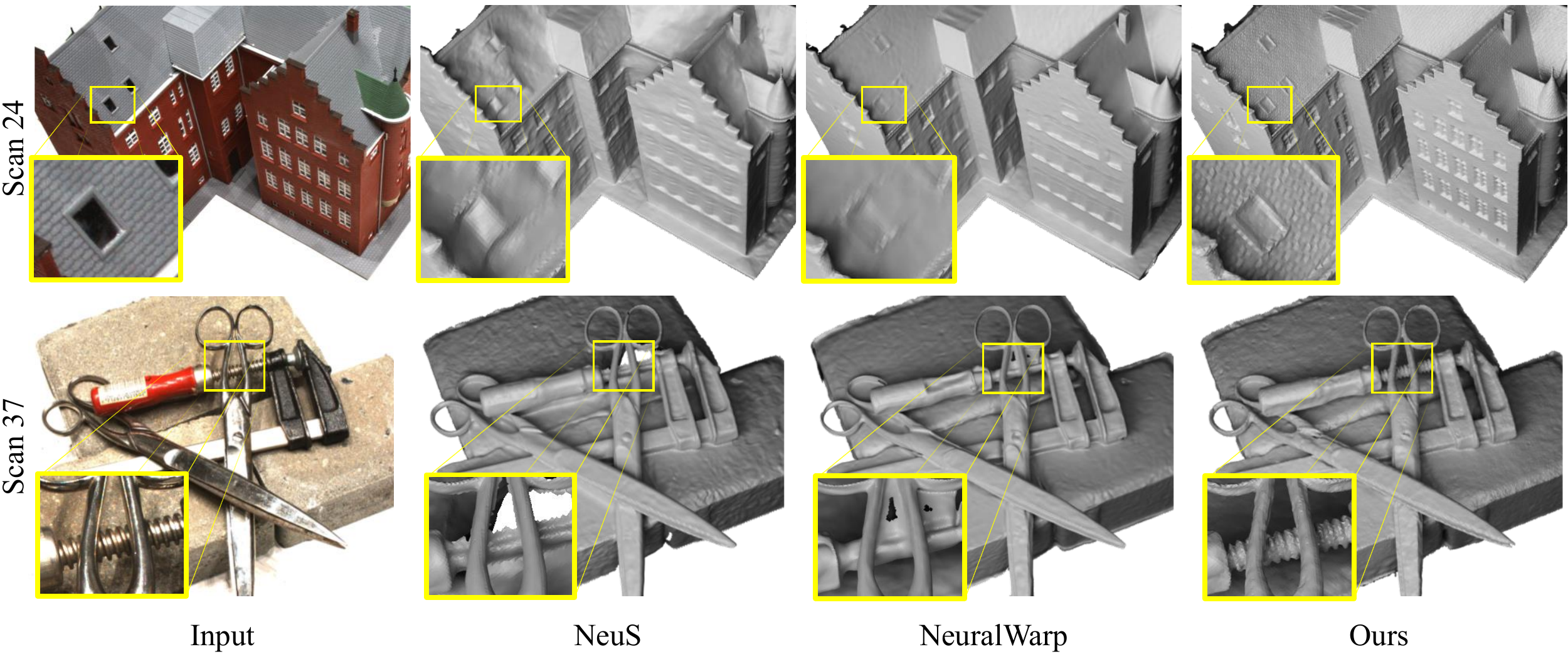}
    \caption{
    \textbf{Qualitative comparison on the DTU benchmark~\cite{jensen2014large}.} 
    \name produces more accurate and higher-fidelity surfaces.}
    \label{fig:dtu}
\end{figure*}

%% file: sections/4-experiments.tex
\input{figures/ablation_c2f.tex}
\input{tables/dtu.tex}

\section{Experiments}

\vspace{4pt}
\noindent\textbf{Datasets.}
Following prior work, we conduct experiments on 15 object-centric scenes of the DTU dataset~\cite{jensen2014large}. 
Each scene has 49 or 64 images captured by a robot-held monocular RGB camera. 
The ground truth is obtained from a structured-light scanner.
We further conduct experiments on 6 scenes of the Tanks and Temples dataset~\cite{knapitsch2017tanks}, including large-scale indoor/outdoor scenes.
Each scene contains 263 to 1107 images captured using a hand-held monocular RGB camera. 
The ground truth is obtained using a LiDAR sensor.

\vspace{4pt}
\noindent\textbf{Implementation details.}
Our hash encoding resolution spans $2^5$ to $2^{11}$ with 16 levels.
Each hash entry has a channel size of 8.
The maximum number of hash entries of each resolution is $2^{22}$.
We activate 4 and 8 hash resolutions at the beginning of optimization for DTU dataset and \tnt{} respectively, due to differences in scene scales.
We enable a new hash resolution every 5000 iterations when the step size $\epsilon$ equals its grid cell size.
For all experiments, we do \textit{not} utilize auxiliary data such as segmentation or depth during the optimization process.

\vspace{4pt}
\noindent\textbf{Evaluation criteria.} We report Chamfer distance and F1 score for surface evaluation~\cite{jensen2014large,knapitsch2017tanks}. 
We use peak signal-to-noise ratio (PSNR) to report image synthesis qualities.

\input{figures/tt.tex}

\subsection{DTU Benchmark}
We show qualitative results in Fig.~\ref{fig:dtu} and quantitative results in Table~\ref{tab:dtu_result}.
On average, \name achieves the lowest Chamfer distance and the highest PSNR, even without using auxiliary inputs.
The result suggests that \name is more generally applicable than prior work when recovering surfaces and synthesizing images, despite not performing best in every individual scene.

We further ablate \name against the following conditions: 1) \ag: analytical gradients, 2) \agl: analytical gradients and progressive activating hash resolutions, 3) \fd: numerical gradients with varying $\epsilon$. 
Fig.~\ref{fig:ablation_c2f} shows the results qualitatively.
\ag produces noisy surfaces, even with hash resolutions progressively activated (\agl).
\fd improves the smoothness of the surface, sacrificing details.
Our setup (\fdl) produces both smooth surfaces and fine details.

\input{figures/lod.tex}
\input{tables/tt.tex}

\subsection{\tnt{}}
As no public result is available for \tnt{}, we train NeuS~\cite{wang2021neus} and NeuralWarp~\cite{darmon2022improving} following our setup.
We also report classical multi-view stereo results using COLMAP~\cite{schoenberger2016sfm}.
As COLMAP and NeuralWarp do not support view synthesis, we only report PSNR from NeuS.
Results are summarized in Fig.~\ref{fig:tt} and Table~\ref{tab:tt_result}.

\name achieves the highest average PSNR and performs best in terms of F1 score. 
Comparing against NeuS~\cite{wang2021neus}, we can recover high-fidelity surfaces with intricate details.
We find that the dense surfaces generated from COLMAP are sensitive to outliers in the sparse point cloud.
We also find that NeuralWarp often predicts surfaces for the sky and backgrounds potentially due to their color rendering scheme following VolSDF~\cite{yariv2021volume}.
The additional surfaces predicted for backgrounds are counted as outliers and worsen F1 scores significantly.
We instead follow NeuS~\cite{wang2021neus} and use an additional network~\cite{zhang2020nerf++} to model the background.

Similar to the DTU results, using the analytical gradient produces noisy surfaces and thus leads to a low F1 score.
We further note that the reconstruction of Courthouse shown in Figs.~\ref{fig:teaser} and~\ref{fig:tt} are the same building of different sides, demonstrating the capability of \name for large-scale granular reconstruction.

\subsection{Level of Details}
As \name progressively optimizes the hash features of increasing resolution, we inspect the progressive level of details similar to NGLOD~\cite{takikawa2021neural}. 
We show a qualitative visualization in Fig.~\ref{fig:lod}.
While some surfaces are entirely missed by coarse levels, for example, the tree, table, and bike rack, these structures are recovered by finer resolutions successfully.
The ability to recover missing surfaces demonstrates the advantages of our spatial hierarchy-free design. 

Moreover, we note that flat surfaces are predicted at sufficiently high resolutions (around Level 8 in this example). 
Thus, only relying on the continuity of local cells of coarse resolutions is not sufficient to reconstruct large continuous surfaces. 
The result motivates the use of the numerical gradients for the higher-order derivatives, such that back-propagation is beyond local grid cells.

\subsection{Ablations}
\input{figures/ablation.tex}

\vspace{4pt}
\noindent
\textbf{Curvature regularization.}
We ablate the necessity of curvature regularization in \name and compare the results in Fig.~\ref{fig:ablation}(a).
Intuitively, $\lcurv$ acts as a smoothness prior by minimizing surface curvatures.
Without $\lcurv$, we find that the surfaces tend to have undesirable sharp transitions.
By using $\lcurv$, the surface noises are removed.

\vspace{4pt}
\noindent
\textbf{Topology warmup.}
We follow prior work and initialize the SDF approximately as a sphere~\cite{yariv2020multiview}. 
With an initial spherical shape, using $\lcurv$ also makes concave shapes difficult to form because $\lcurv$ preserves topology by preventing singularities in curvature. 
Thus, instead of applying $\lcurv$ from the beginning of the optimization process, we use a short warmup period that linearly increases the curvature loss strength. 
We find this strategy particularly helpful for concave regions, as shown in Fig.~\ref{fig:ablation}(b).

%% file: figures/ablation_c2f.tex
\begin{figure*}[bt]
    \centering
    \includegraphics[width=\linewidth]{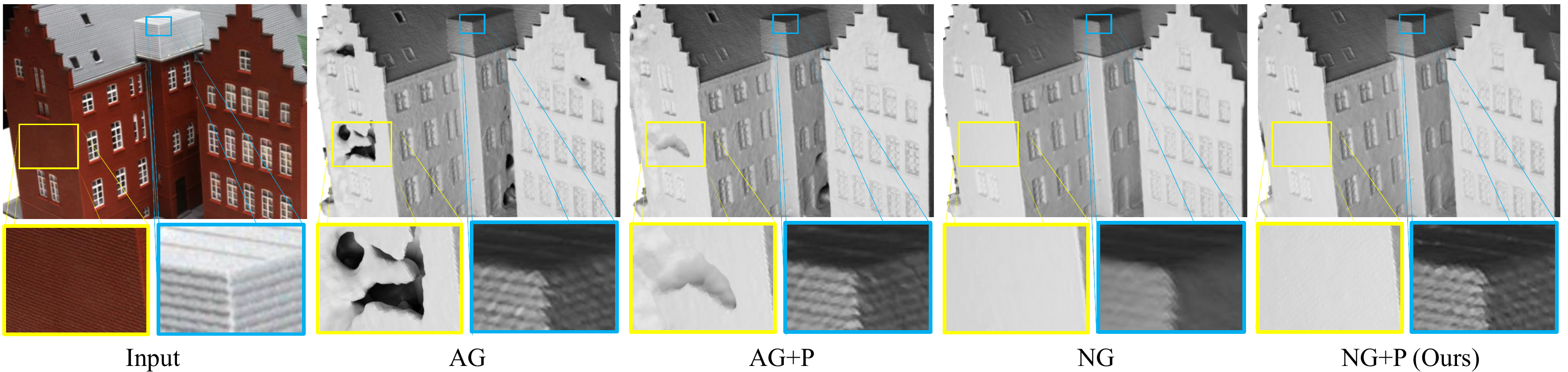}
    \caption{\textbf{Qualitative comparison of different coarse-to-fine optimization scheme.} When using the analytical gradient (\ag and \agl), coarse surfaces often contain artifacts. While using numerical gradients (\fd) leads to a better coarse shape, details are also smoothed. Our solution (\fdl) produces both smooth surfaces and fine details.
    }
    \label{fig:ablation_c2f}
\end{figure*}

%% file: tables/dtu.tex
\setlength\tabcolsep{0.5em}
\begin{table*}[tb]
\centering
\ra{1.1}
\resizebox{\textwidth}{!}{%
\begin{tabular}{@{}llcccccccccccccccccc}
\specialrule{.2em}{.1em}{.1em}
 \multicolumn{3}{c}{} & 24 & 37 & 40 & 55 & 63 & 65 & 69 & 83 & 97 & 105 & 106 & 110 & 114 & 118 & 122 & & Mean \\ \cline{4-18} \cline{20-20}
\multirow{10}{*}{\rotatebox[origin=c]{90}{Chamfer distance (mm) $\downarrow$}} & NeRF\cite{mildenhall2020nerf} & & 1.90 & 1.60 & 1.85 & 0.58 & 2.28 & 1.27 & 1.47 & 1.67 & 2.05 & 1.07 & 0.88 & 2.53 & 1.06 & 1.15 & 0.96 & & 1.49 \\
 & VolSDF\cite{yariv2021volume} & & 1.14 & 1.26 & 0.81 & 0.49 & 1.25 & 0.70 & 0.72 & 1.29 & 1.18 & 0.70 & 0.66 & 1.08 & 0.42 & 0.61 & 0.55 & & 0.86 \\
 & NeuS\cite{wang2021neus} & & 1.00 & 1.37 & 0.93 & 0.43 & 1.10 & 0.65 & \sbest 0.57 & 1.48 & 1.09 & 0.83 & 0.52 & 1.20 & 0.35 & 0.49 & 0.54 & & 0.84 \\
 & HF-NeuS\cite{wanghf} & & 0.76 & 1.32 & 0.70 & 0.39 & 1.06 & 0.63 & 0.63 & \sbest 1.15 & 1.12 & 0.80 & 0.52 & 1.22 & 0.33 & 0.49 & 0.50 & & 0.77 \\
 & RegSDF\cite{zhang2022critical} $\dagger$ & & 0.60 & 1.41 & 0.64 & 0.43 & 1.34 & \sbest 0.62 & 0.60 & \best 0.90 & \best 0.92 & 1.02 & 0.60 & \best 0.59 & \best 0.30 & \best 0.41 & \best 0.39 & & 0.72 \\
 & NeuralWarp\cite{darmon2022improving}  $\dagger$ & & 0.49 & \best 0.71 & \sbest 0.38 & 0.38 & \best 0.79 & 0.81 & 0.82 & 1.20 & 1.06 & \best 0.68 & 0.66 & \sbest 0.74 & 0.41 & 0.63 & 0.51 & & 0.68 \\
 \cline{2-2} \cline{4-18} \cline{20-20}
 & \ag & & 0.67 & 1.04 & 0.84 & 0.39 & 1.43 & 1.23 & 1.11 & 1.24 & 1.54 & 0.85 & 0.50 & 1.01 & 0.37 & 0.51 & 0.44 & & 0.88 \\ 
 & \agl & & 0.59 & 0.95 & 0.46 & \best 0.34 & 1.19 & 0.70 & 0.79 & 1.19 & 1.37 & \sbest 0.69 & \sbest 0.49 & 0.93 & 0.33 & \sbest 0.44 & 0.44 & & 0.73 \\
 & \fd & & \sbest 0.48 & 0.81 & 0.43 & \sbest 0.35 & 0.89 & 0.71 & 0.61 & 1.26 & 1.06 & 0.74 & \best 0.47 & 0.79 & 0.33 & 0.45 & \sbest  0.43 & & \sbest 0.65 \\
 & \fdl (Ours) & & \best 0.37 & \sbest 0.72 & \best \best 0.35 & \sbest 0.35 & \sbest 0.87 & \best 0.54 & \best 0.53 & 1.29 & \sbest 0.97 & 0.73 & \best 0.47 & \sbest 0.74 & \sbest 0.32 & \best 0.41 & \sbest  0.43 & & \best 0.61 \\
 \specialrule{.13em}{.1em}{.1em}
\multirow{8}{*}{\rotatebox[origin=c]{90}{PSNR $\uparrow$}} 
 & RegSDF\cite{zhang2022critical} $\dagger$ & & 24.78 & 23.06 & 23.47 & 22.21 & 28.57 & 25.53 & 21.81 & 28.89 & 26.81 & 27.91 & 24.71 & 25.13 & 26.84 & 21.67 & 28.25 & & 25.31 \\
 & NeuS\cite{wang2021neus} & & 26.62 & 23.64 & 26.43 & 25.59 & 30.61 & \sbest 32.83 & 29.24 & 33.71 & 26.85 & 31.97 & 32.18 & 28.92 & 28.41 & 35.00 & 34.81 & & 29.79 \\
 & VolSDF\cite{yariv2021volume} & & 26.28 & 25.61 & 26.55 & 26.76 & 31.57 & 31.50 & 29.38 & 33.23 & 28.03 & 32.13 & 33.16 & 31.49 & 30.33 & 34.90 & 34.75 & & 30.38 \\
 & NeRF\cite{mildenhall2020nerf} & & 26.24 & \sbest 25.74 & 26.79 & 27.57 & \sbest 31.96 & 31.50 & 29.58 & 32.78 & \sbest 28.35 & 32.08 & 33.49 & \sbest 31.54 & \sbest 31.00 & 35.59 & 35.51 & & 30.65 \\
  \cline{2-2} \cline{4-18} \cline{20-20}
 & \ag & & 29.97 & 24.98 & 23.11 & 30.27 & 30.60 & 31.27 & 29.27 & 34.22 & 27.47 & 33.09 & 33.85 & 29.98 & 29.41 & 35.69 & 35.11 & & 30.55 \\
 & \agl & & 30.12 & 24.63 & 29.59 & 30.29 & 31.60 & 32.04 & \sbest 29.85 & 34.19 & 27.82 & 33.23 & 33.95 & 29.15 & 29.44 & 35.99 & 35.67 & & 31.17 \\
 & \fd &  & \sbest 30.34 & 25.14 & \sbest 30.20 & \sbest 30.79 & 31.72 & 31.86 & 29.81 & \sbest 34.36 & 28.01 & \sbest 33.45 & \sbest 34.38 & 30.39 & 29.88 & \sbest 36.02 & \sbest 35.74 & & \sbest 31.47 \\
 & \fdl (Ours) & &  \best 30.64 & \best 27.78 & \best 32.70 & \best 34.18 & \best 35.15 & \best 35.89 & \best 31.47 & \best 36.82 & \best 30.13 & \best 35.92 & \best 36.61 & \best 32.60 & \best 31.20 & \best 38.41 & \best 38.05 & & \best 33.84 \\
 \specialrule{.2em}{.1em}{.1em}
\end{tabular}%
}
\caption{\textbf{Quantitative results on DTU dataset~\cite{jensen2014large}.} 
\name achieves the best reconstruction accuracy and image synthesis quality.
{\best Best result}. {\sbest Second best result}.
$\dagger$~Requires 3D points from \SFM.
Best viewed in color.
}
\label{tab:dtu_result}
\end{table*}

%% file: figures/tt.tex
\begin{figure*}[tb]
    \centering
    \includegraphics[width=\linewidth]{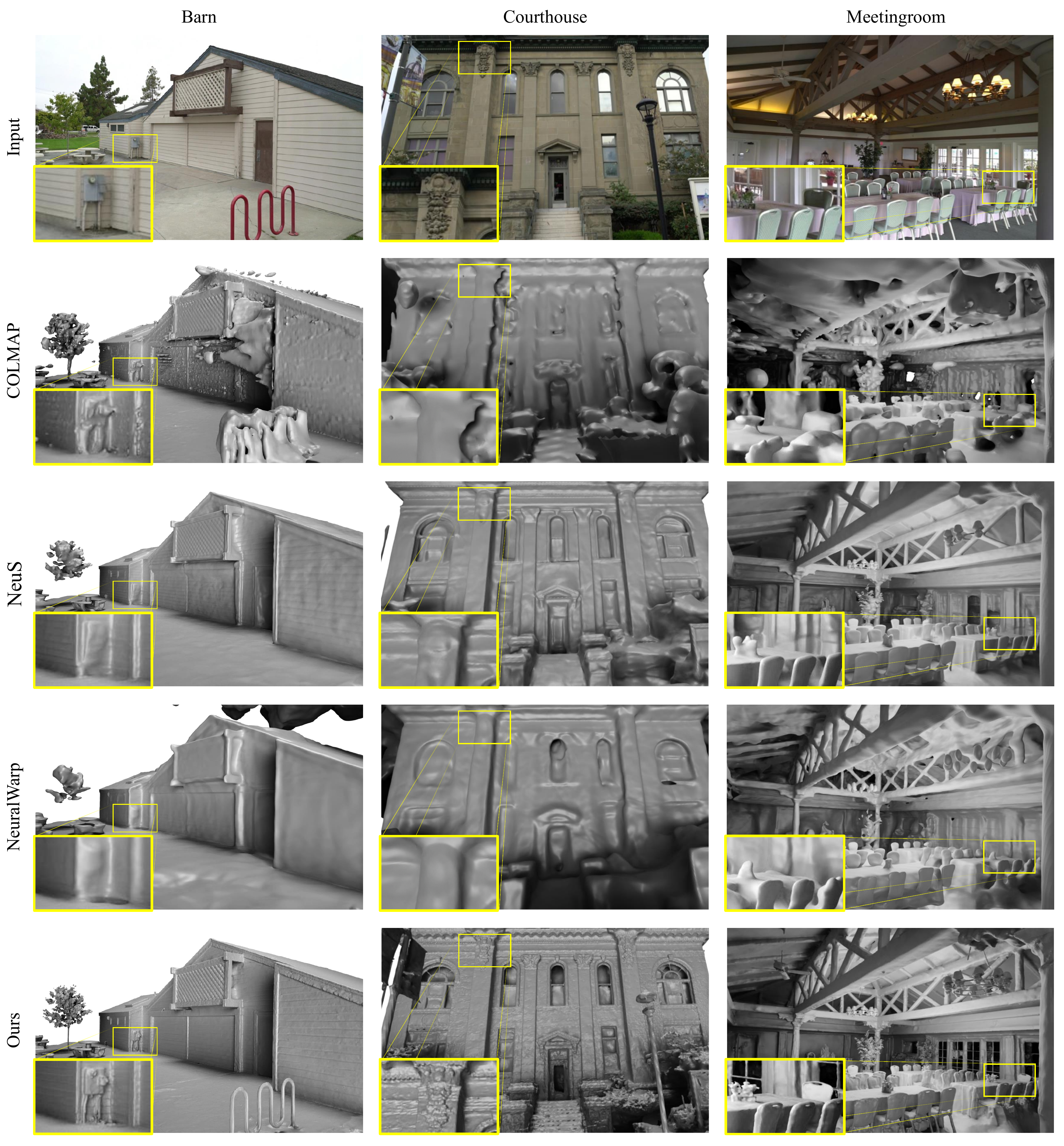}
    \caption{
    \textbf{Qualitative comparison on \tnt dataset~\cite{knapitsch2017tanks}.}
    \name captures the scene details better compared to other baseline approaches, while baseline approaches have missing or noisy surfaces.
    }
    \vspace{-4pt}
    \label{fig:tt}
\end{figure*}

%% file: figures/lod.tex
\begin{figure*}[bt]
    \centering
    \includegraphics[width=\linewidth,page=1]{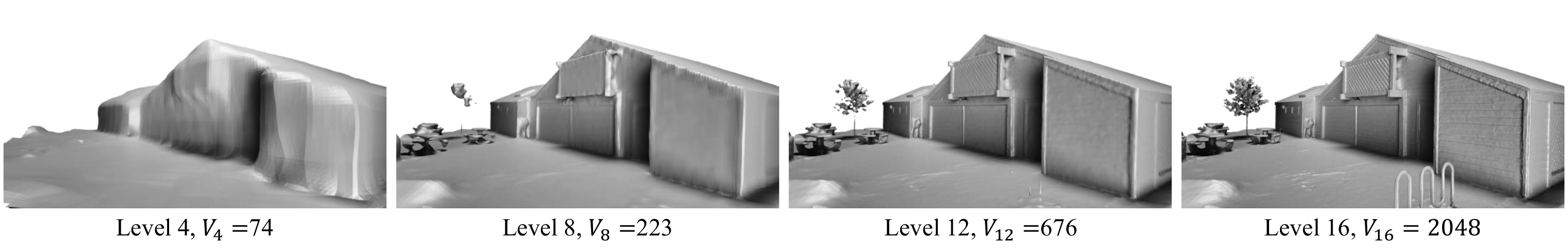}
    \caption{
    \textbf{Results at different hash resolutions.} 
    While some structures, such as the tree, table, and bike rack, are missed at coarse resolutions (Level 4). 
    Finer resolutions can progressively recover these missing surfaces.
    Flat continuous surfaces also require sufficiently fine resolutions to predict (Level 8).
    The result motivates the non-local updates when using numerical gradients for higher-order derivatives.
    }
    \label{fig:lod}
\end{figure*}

%% file: tables/tt.tex
\setlength\tabcolsep{1em}
\begin{table*}[tb]
\centering
\ra{1.1}
\resizebox{\textwidth}{!}{%
\begin{tabular}{@{}lccccccccccccc}
\specialrule{.2em}{.1em}{.1em}
 & \multicolumn{7}{c}{F1 Score $\uparrow$} & & \multicolumn{5}{c}{PSNR $\uparrow$} \\ \cline{2-8} \cline{10-14}
 & \multirow{2}{*}{\shortstack{NeuralWarp \\ \cite{darmon2022improving}}} & \multirow{2}{*}{\shortstack{COLMAP \\ \cite{schoenberger2016sfm}}} & \multirow{2}{*}{\shortstack{NeuS \\ \cite{wang2021neus}}} & \multirow{2}{*}{\ag} & \multirow{2}{*}{\agl} & \multirow{2}{*}{\fd} & \multirow{2}{*}{\shortstack{\fdl \\ (Ours)}} & & \multirow{2}{*}{\shortstack{NeuS \\ \cite{wang2021neus}}} & \multirow{2}{*}{\ag} & \multirow{2}{*}{\agl} & \multirow{2}{*}{\fd} & \multirow{2}{*}{\shortstack{\fdl \\ (Ours)}} \\ 
 & & & & & & & & & & & & & \\ \specialrule{.13em}{.1em}{.1em}
Barn & 0.22 & 0.55 & 0.29 & 0.22 & 0.31 & \sbest 0.63 & \best 0.70 & & 26.36 &\sbest 26.91 & 26.69 & 26.14 & \best 28.57 \\
Caterpillar & 0.18 & 0.01 & 0.29 & 0.23 & 0.24 & \sbest 0.30 & \best 0.36 & & 25.21 & \sbest 26.04 & 25.12 & 26.16 & \best 27.81 \\
Courthouse & 0.08 & 0.11 & 0.17 & 0.08 & 0.09 & \sbest 0.24 & \best 0.28 & & 23.55 & 25.43 & \sbest 25.63 & 25.06 & \best 27.23 \\
Ignatius & 0.02 & 0.22 & 0.83 & 0.72 & 0.73 & \sbest 0.85 & \best 0.89 & & \sbest 23.27 & 22.69 & 22.73 & \best 23.78 & \sbest 23.67 \\
Meetingroom & 0.08 & 0.19 & 0.24 &  0.04 & 0.05 & \sbest 0.27 & \best 0.32 & & 25.38 & \sbest 28.13 & 28.05 & 27.44 & \best 30.70 \\
Truck & 0.35 & 0.19 & \sbest 0.45 & 0.33 & 0.37 & 0.44 & \best 0.48 & & 23.71 & 23.89 & \sbest 23.95 & 22.99 & \best 25.43 \\ \hline
Mean & 0.15 & 0.21 & 0.38 & 0.27 & 0.30 & \sbest 0.45 & \best 0.50 & & 24.58 & \sbest 25.51 & 25.36 & 25.26 & \best 27.24 \\
\specialrule{.2em}{.1em}{.1em}
\end{tabular}%
}
\caption{
\textbf{Quantitative results on \tnt dataset~\cite{knapitsch2017tanks}.} 
\name achieves the best surface reconstruction quality and performs best on average in terms of image synthesis.
{\best Best result}. {\sbest Second best result.}
Best viewed in color.
}
\label{tab:tt_result}
\end{table*}

%% file: figures/ablation.tex
\begin{figure}[bt]
    \centering
    \includegraphics[width=\linewidth]{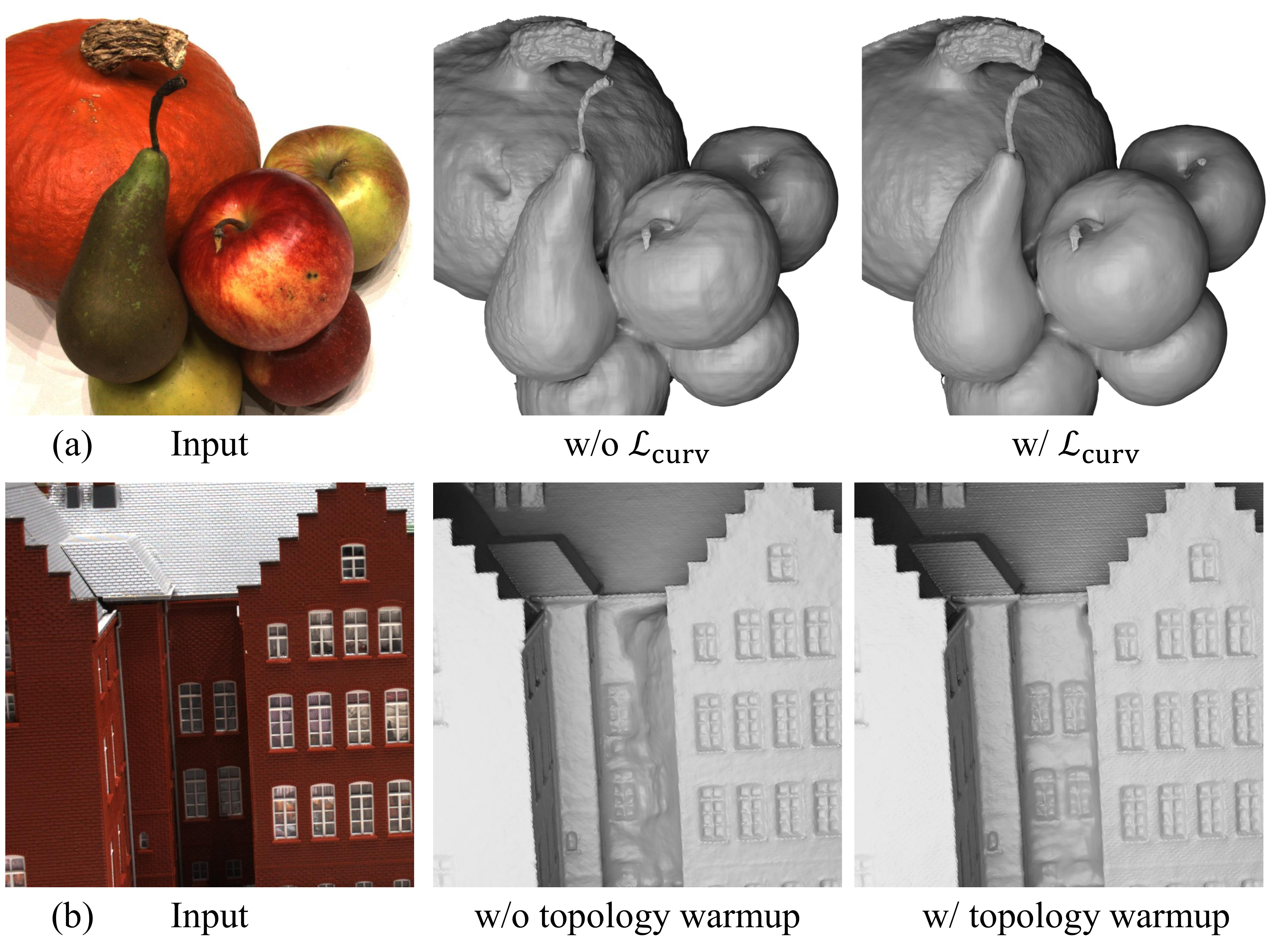}
    \caption{\textbf{Ablation results.}
    (a) Surface smoothness improves with curvature regularization $\lcurv$. (b) Concave shapes are better formed with topology warmup. }
    \label{fig:ablation}
\end{figure}

%% file: sections/5-conclusion.tex
\section{Conclusion}

We introduce \name, an approach for photogrammetric neural surface reconstruction. 
The findings of \name are simple yet effective: using numerical gradients for higher-order derivatives and a coarse-to-fine optimization strategy. 
\name unlocks the representation power of multi-resolution hash encoding for neural surface reconstruction modeled as SDF. 
We show that \name effectively recovers dense scene structures of both object-centric captures and large-scale indoor/outdoor scenes with extremely high fidelity, enabling detailed large-scale scene reconstruction from RGB videos.
Our method currently samples pixels from images randomly without tracking their statistics and errors.
Therefore, we use long training iterations to reduce the stochastics and ensure sufficient sampling of details.
It is our future work to explore a more efficient sampling strategy to accelerate the training process.

%% file: sections/6-appendix.tex
\noindent\textbf{Acknowledgements.}
We thank Alexander Keller, Tsung-Yi Lin, Yen-Chen Lin, Stan Birchfield, Zan Gojcic, Tianchang Shen, and Zian Wang for helpful discussions and paper proofreading.
This work was done during Zhaoshuo Li's internship at NVIDIA Research and funded in part by NIDCD K08 Grant DC019708.

\appendix
\section{Additional Hyper-parameters}
Following prior work~\cite{wang2021neus,yariv2020multiview,yariv2021volume}, we assume the region of interest is inside a unit sphere.
The total number of training iterations is 500k.
When a given hash resolution is not active, we set the feature vectors to zero.
We use a learning rate of $1\times10^{-3}$ with a linear warmup of 5k iterations.
We decay the learning rate by a factor of 10 at 300k and 400k.
We use AdamW~\cite{loshchilov2017decoupled} optimizer with a weight decay of $10^{-2}$.
We set $ w_\text{eik}=0.1$.
The curvature regularization strength $w_{curv}$ linearly warms up $5\times10^{-4}$ following the schedule of learning rate and decays by the same spacing factor between hash resolutions every time $\epsilon$ decreases.
The SDF MLP has one layer, while the color MLP has four layers.
For the DTU benchmark, we follow prior work~\cite{yariv2020multiview,wang2021neus,yariv2021volume} and use a batch size of 1.
For the \tnt{} dataset, we use a batch size of 16.
We use the marching cubes algorithm~\cite{lorensen1987marching} to convert predicted SDF to triangular meshes.
The marching cubes resolution is set to 512 for the DTU benchmark following prior work~\cite{yariv2020multiview,wang2021neus,yariv2021volume,darmon2022improving} and 2048 for the \tnt dataset.

\section{Additional In-the-wild Results}
We present additional in-the-wild results collected at the NVIDIA HQ Park and Johns Hopkins University in \autoref{fig:app_nvidia_jhu}. 
The videos are captured by a consumer drone.
The camera intrinsics and poses are recovered using COLMAP~\cite{schoenberger2016sfm}.
To define the bounding regions, we have developed an open-sourced Blender add-on\footnote{\selectfont \url{https://github.com/mli0603/BlenderNeuralangelo}} to allow users interactively select regions of interest using the sparse point cloud from COLMAP.
The surfaces are reconstructed using the same setup and hyperparameters as the \tnt dataset.
\name successfully reconstructs complex geometries and scene details, such as the buildings, sculptures, trees, umbrellas, walkways, and \etc.
Using the same setup as \tnt also suggests that \name is generalizable with the proposed set of hyper-parameters.
\input{figures/appendix_nvidia_jhu}

\section{Additional \tnt Results}
We present additional results on the \tnt dataset~\cite{knapitsch2017tanks} in this section.

\vspace{4pt}
\noindent
\textbf{Surface reconstruction.}
Concurrent with our work, Geo-NeuS~\cite{fu2022geo} uses the sparse point clouds from COLMAP~\cite{schoenberger2016sfm} to improve the surface quality.
However, we find that in large-scale in-the-wild scenes, the COLMAP point clouds are often noisy, even after filtering.
Using the noisy point clouds may degrade the results, similarly observed in~\cite{zhang2022critical}.
As evidence, we benchmark Geo-NeuS~\cite{fu2022geo} on \tnt (Table~\ref{tab:app_tt}).
We find that Geo-NeuS performs worse than NeuS and \name in most scenes.

\input{tables/app_tt.tex}

\vspace{4pt}
\noindent
\textbf{RGB image synthesis.}
Due to similarities between adjacent video frames, we report PSNR by sub-sampling 10 times input video temporally and evaluating the sub-sampled video frames. Qualitative comparison of \name and prior work NeuS~\cite{wang2021neus} is shown in Fig~\ref{fig:app_tt_rendering}. \name produces high-fidelity renderings compared to NeuS~\cite{wang2021neus}, with details on the buildings and objects recovered. 
Neither COLMAP~\cite{schoenberger2016sfm} nor NeuralWarp~\cite{darmon2022improving} supports view synthesis or accounts for view-dependent effects. 
Thus, we only report the F1 score of the reconstructed surfaces for these two approaches.

\input{figures/appendix_rendering_tt.tex}
\input{figures/appendix_additional_dtu.tex}
\input{figures/appendix_rendering_dtu.tex}
\input{tables/dtu_mask.tex}

\section{Additional DTU Results}
We present additional results on the DTU benchmark~\cite{jensen2014large} in this section.

\vspace{4pt}
\noindent
\textbf{Surface reconstruction.}
We visualize the reconstructed surfaces of additional scenes of the DTU benchmark. Qualitative comparison with NeuS~\cite{wang2021neus} and NeuralWarp~\cite{darmon2022improving} are shown in Fig.~\ref{fig:app_dtu_additional_results}.

Compared to prior work, \name not only can reconstruct smoother surfaces such as in Scan 40, 63, and 69 but also produces sharper details such as in Scan 63 and 118 (\eg the details of the pumpkin vine and the statue face).
While \name performs better on average across scenes, we note that the qualitative result of \name does not improve significantly in Scan 122, where the object of interest has mostly diffuse materials and relatively simple textures.
Moreover, we find that \name fails to recover details compared to NeuS~\cite{wang2021neus} when the scene is highly reflective, such as Scan 69.
\name misses the button structures and eyes.
Such a finding agrees with the results of Instant NGP~\cite{muller2022instant}, where NeRF using Fourier frequency encoding and deep MLP performs favorably against multi-resolution hash encoding for highly reflective surfaces. 
Future work on improving the robustness of \name in reflective scenes, a drawback inherited from hash encoding, can further generalize the application of \name. 

\input{figures/appendix_dtu_mask.tex}

\vspace{4pt}
\noindent
\textbf{RGB image synthesis.}
In the paper, we report the PSNR result of \name to quantify the image synthesis quality. 
Due to the simplicity of the background, we only evaluate the PSNR of the foreground objects given the object masks.
We visualize the rendered images in Fig.~\ref{fig:app_dtu_rendering}.
We only choose NeuS~\cite{wang2021neus} as our baseline as NeuralWarp~\cite{darmon2022improving} does not generate rendered images.

Fig.~\ref{fig:app_dtu_rendering} shows that \name successfully renders the detailed textures while NeuS produces overly smoothed images. The results suggest that \name is able to produce high-fidelity renderings and capture details better.

\vspace{4pt}
\noindent
\textbf{DTU foreground mask.}
The foreground object masks are used to remove the background for proper evaluation~\cite{yariv2020multiview,wang2021neus,zhang2021learning,oechsle2021unisurf,darmon2022improving} on the DTU benchmark. 
We follow the evaluation protocol of NeuralWarp~\cite{darmon2022improving} and dilate the object masks by 12 pixels.
In all prior work, the foreground object masks used are annotated and provided by the authors of IDR~\cite{yariv2020multiview}. 
However, we find that the provided masks are imperfect in Scan 83. 
Fig.~\ref{fig:dtu_mask} shows that part of the object is annotated as background. 
The masks provided by IDR also only include the foreground objects while the ground truth point clouds include the brick holding the objects. 
Thus, we manually annotate Scan 83 and report the updated results in Table~\ref{tab:dtu_mask} for additional comparison.
We note that fixing the object masks for Scan 83 leads to improved results across all methods. 

\section{Additional Ablations}
We conduct additional ablations and summarize the results in this section.

\vspace{4pt}
\noindent \textbf{Color network.}
For the \tnt dataset, we add per-image latent embedding to the color network following NeRF-W~\cite{martin2021nerf} to model the exposure variation across frames.
Qualitative results are shown in Fig.~\ref{fig:app_per_img_embed}. 
After introducing the per-image embedding, the floating objects used to explain exposure variation have been greatly reduced.
\input{figures/appendix_per_img_embed}

\vspace{4pt}
\noindent
\textbf{Curvature regularization strength.}
The curvature regularization adds a smoothness prior to the optimization.
As the step size $\epsilon$ decreases and finer hash grids are activated, finer details may be smoothed if the curvature regularization is too strong.
To avoid loss of details, we scale down the curvature regularization strength by the spacing factor between hash resolutions each time the step size $\epsilon$ decreases. 
Details are better preserved by decaying $w_\text{curv}$ (Fig.~\ref{fig:app_curv_decay}).

\input{figures/appendix_curv_decay}
\input{figures/appendix_normal_comparison}
\input{figures/appendix_color_network}
\input{figures/appendix_color_result}

\vspace{4pt}
\noindent
\textbf{Numerical \textit{v.s} analytical gradient.}
We visualize in Fig.~\ref{fig:app_normal_comparison} the surface normals computed by using both numerical and analytical gradients after the optimization finishes.
At the end of the optimization, the step size $\epsilon$ has decreased sufficiently small to the grid size of the finest hash resolution. 
Using numerical gradients is nearly identical to using analytical gradients.
Fig.~\ref{fig:app_normal_comparison} shows that the surface normals computed from both numerical and analytical gradients are indeed qualitatively similar, with negligible errors scattered across the object.

\vspace{4pt}
\noindent
\textbf{Color network.}
By default, we follow prior work~\cite{yariv2020multiview,wang2021neus} and predict color conditioned on view direction, surface normal, point location, and features from the SDF MLP.
We use spherical harmonics following~\cite{yu2021plenoxels} to encode view direction as it provides meaningful interpolation in the angular domain.
When the data is captured with exposure variation in the wild, such as the \tnt dataset, we further add per-image appearance encoding following NeRF-W~\cite{martin2021nerf}.

We have also implemented a more explicit color modeling process.
The color network is shown in Fig.~\ref{fig:intrinsic_decomp}, attempting to better disentangle color-shape ambiguities. 
However, we do \textit{not} observe improvements in surface qualities using such a decomposition design.
The intrinsic decomposed color network contains two branches -- albedo and shading branches.
The final rendered image $C \in \mathbb{R}^3$ is the sum of the albedo image $C_a$ and shading image $C_s$:
\begin{equation}
    C = \Phi(C_a + C_s),
\end{equation}
where $\Phi$ is the Sigmoid function to normalize the predictions into the range of 0 to 1.

The albedo branch predicts RGB values $C_a \in \mathbb{R}^3$ that are view-invariant.
It receives point locations and features from the SDF MLP as input.
On the other hand, the shading branch predicts gray values $C_s \in \mathbb{R}$ that is view dependent to capture reflection, varying shadow, and exposure changes.
We opt for the single channel design for the shading branch as specular highlights, exposure variations, and moving shadows are often intensity changes~\cite{rudnev2022nerf}.
The single-channel gray color design also encourages the albedo branch to learn the view-invariant color better as the shading branch is limited in its capacity.
Other than the point locations and SDF MLP features, the shading branch is additionally conditioned on reflection direction and view direction following RefNeRF~\cite{verbin2022ref} to encourage better shape recovery.
We use two hidden layers for the albedo branch and two hidden layers for the diffuse branch to make a fair comparison with the default color network proposed by IDR~\cite{yariv2020multiview}.

We find that with the decomposed color network, the shading branch indeed successfully explains view-dependent effects (Fig.~\ref{fig:intrinsic_decomp}). However, flat surfaces tend to be carved away, potentially due to the instability of dot product from reflection computation (Fig.~\ref{fig:color_result}). Our future work will explore more principled ways for intrinsic color decomposition.

\vspace{4pt}
\noindent
\textbf{Computation time.}
We compare the training and inference time in Table~\ref{tab:appendix_speed} across different setups using our implementation in PyTorch. 
The experiments are conducted on NVIDIA V100 GPUs. 
We note that the training time per iteration when using numerical gradients is longer than using analytical gradients due to additional queries of SDF.
Using numerical gradients experiences approximately a 1.2 times slowdown compared to using analytical gradients.
As NeuS uses 8-layer MLP for SDF MLP and \name uses 1-layer MLP, using numerical gradients is still faster than NeuS~\cite{wang2021neus}.
We also compare the inference time for surface extraction of $128^3$ resolution.
As numerical gradients are used only for training, the speed for NG and AG are the same.
NG and AG are more than 2 times faster than NeuS~\cite{wang2021neus} due to the shallow MLP.

\input{tables/speed.tex}

\section{Derivation of Frequency Encoding}
In the paper, we show that using analytical gradients for higher-order derivatives of multi-resolution hash encoding suffers from gradient locality. 
We show in this section that Fourier frequency encoding~\cite{tancik2020fourier}, which empowers prior work~\cite{yariv2020multiview,yariv2021volume,wang2021neus} on neural surface reconstruction, does not suffer from such locality issue.

Given a 3D position $\pos_{i}$, let the $l$-th Fourier frequency encoding be
\begin{equation}
    \feat_l(\pos_{i}) = \big(\text{sin}(2^l\pi\pos_i),\text{cos}(2^l\pi\pos_i) \big) .
\end{equation}
The derivative of $\feat_l(\pos_i)$ \wrt position can thus be calculated as
\begin{equation}
    \deriv{\feat_l(\pos_i)}{\pos_i} = \big( 2^l \pi \cdot \text{cos}(2^l \pi \pos_i), -2^l \pi \cdot \text{sin}(2^l \pi \pos_i)\big).
\end{equation}
We note that $\deriv{\feat_l(\pos_i)}{\pos_i}$ is continuous across the space, and thus does not suffer from the gradient locality issue as the multi-resolution hash encoding. Moreover, the position $\pos_i$ is present in the derivative, thus allowing for second-order derivatives computation \wrt position for the curvature regularization. 

While Fourier frequencies encoding is continuous, our coarse-to-fine optimization with varying step size in theory still anneals over the different frequencies when computing higher-order derivatives for more robust optimization.
We experiment this idea on the DTU benchmark~\cite{jensen2014large} and observed an improved Chamfer distance: from 0.84 to 0.79.
The improvement in surface reconstruction confirms the benefits of using a coarse-to-fine optimization framework.

%% file: figures/appendix_nvidia_jhu.tex
\begin{figure*}[tb]
    \centering
    \includegraphics[width=\linewidth,page=1]{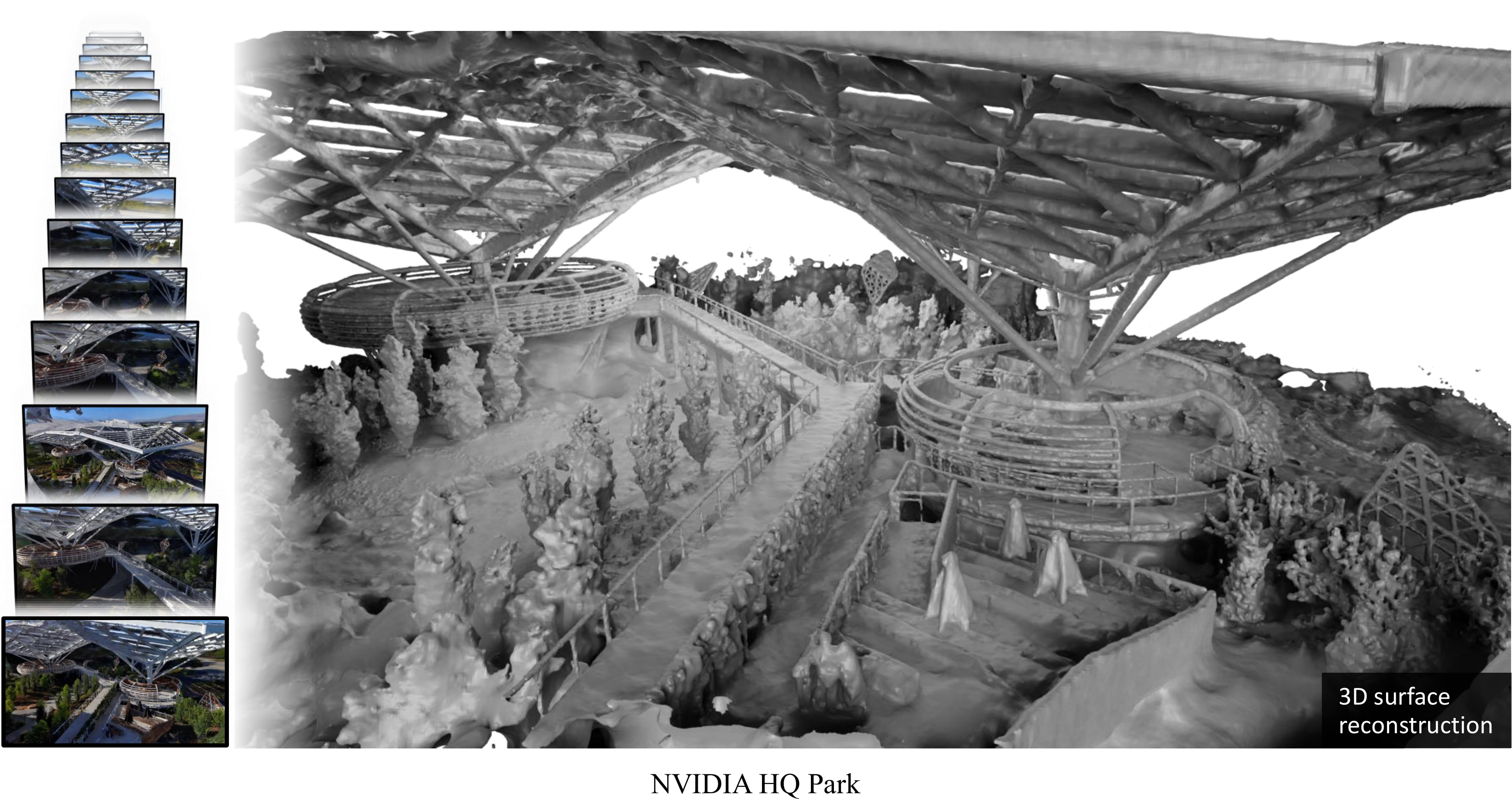}
    \includegraphics[width=\linewidth,page=2]{figures/appendix/appendix_nvidia_jhu.pdf}
    
    \caption{
    Reconstruction results of \textbf{NVIDIA HQ Park} and \textbf{Johns Hopkins University}. Videos are captured by a consumer drone.
    }
    \label{fig:app_nvidia_jhu}
\end{figure*}

%% file: tables/app_tt.tex
\setlength\tabcolsep{1.6em}
\begin{table}[tb]
\centering
\ra{1.1}
\resizebox{\linewidth}{!}{%
\begin{tabular}{@{}lccc}
\specialrule{.2em}{.1em}{.1em}
 & \multicolumn{3}{c@{}}{F1 Score $\uparrow$} \\ \cline{2-4}
 & NeuS~\cite{wang2021neus} & Geo-Neus~\cite{fu2022geo} & Ours \\ \specialrule{.13em}{.1em}{.1em}
Barn & 0.29 & \sbest 0.33 & \best 0.70 \\
Caterpillar & \sbest 0.29 & 0.26 & \best 0.36 \\
Courthouse & \sbest 0.17 & 0.12 & \best 0.28  \\
Ignatius & \sbest  0.83 & 0.72 & \best 0.89 \\
Meetingroom & \sbest  0.24 & 0.20 & \best 0.32  \\
Truck & \sbest 0.45 & \sbest 0.45 & \best 0.48 \\ \hline
Mean & \sbest 0.38 & 0.35 & \best 0.50 \\
\specialrule{.2em}{.1em}{.1em}
\end{tabular}%
}
\caption{
\textbf{Additional quantitative results on \tnt dataset~\cite{knapitsch2017tanks}.} 
\name achieves the best surface reconstruction quality and performs best on average in terms of image synthesis.
{\best Best result}. {\sbest Second best result.}
Best viewed in color.
}
\label{tab:app_tt}
\end{table}

%% file: figures/appendix_rendering_tt.tex
\begin{figure*}[t]
    \centering
    \includegraphics[width=0.98\linewidth]{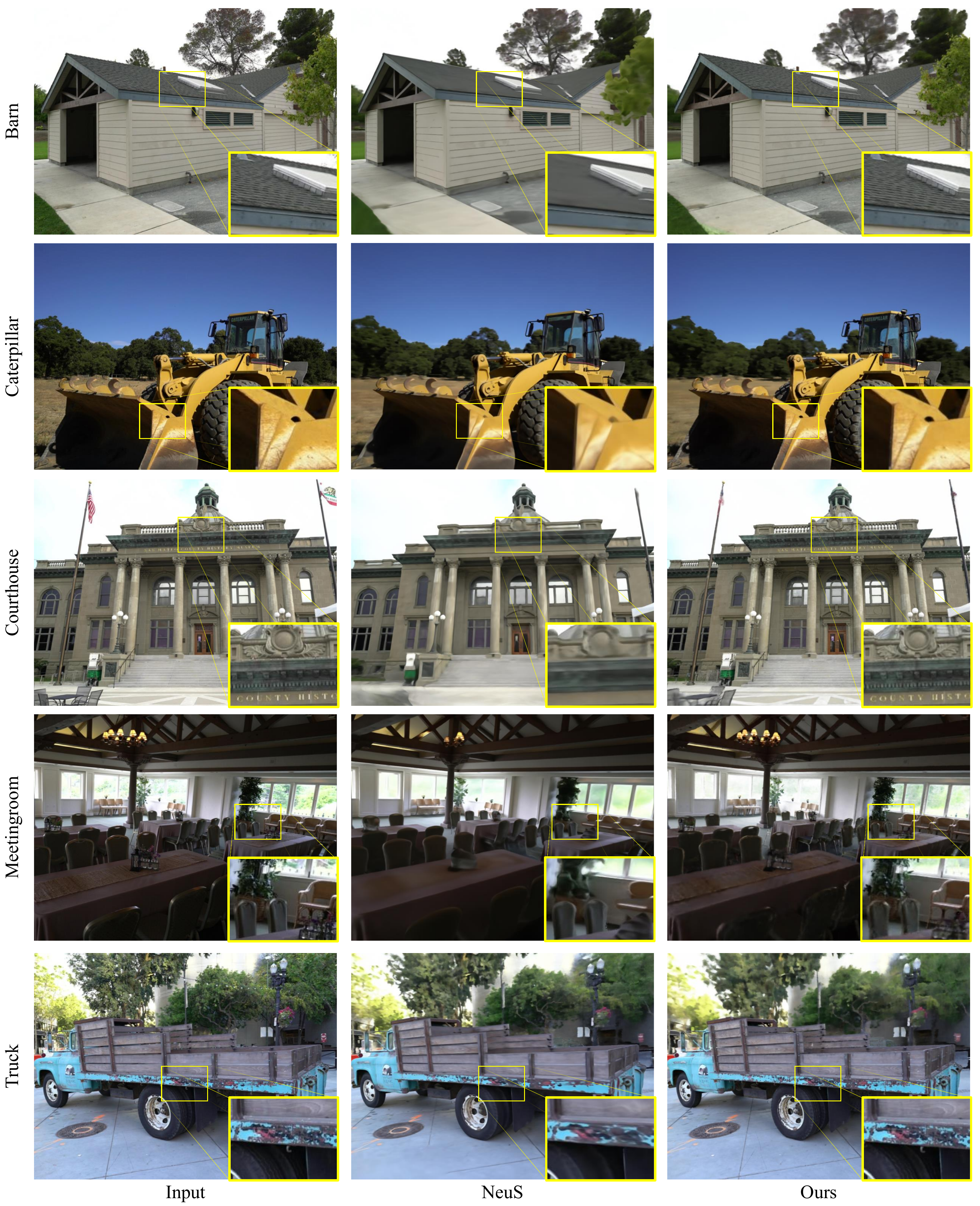}
    \caption{
    \textbf{Qualitative comparison of image rendering on the \tnt dataset~\cite{knapitsch2017tanks}.} 
    Compared to NeuS~\cite{wang2021neus}, \name generates high-quality renderings with texture details on the buildings and objects.
    }
    \label{fig:app_tt_rendering}
\end{figure*}

%% file: figures/appendix_additional_dtu.tex
\begin{figure*}[tb]
    \centering
    \includegraphics[width=\linewidth]{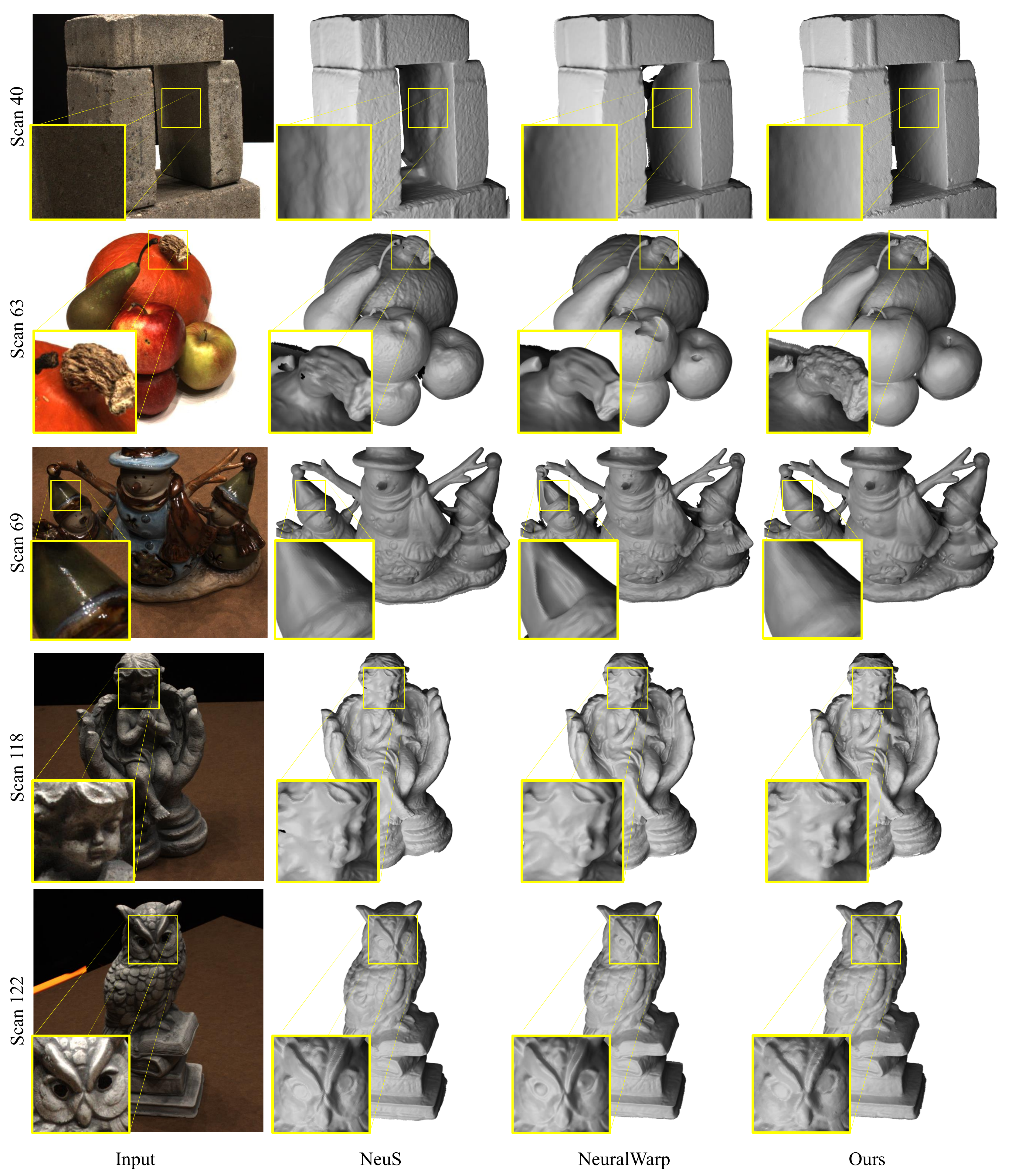}
    \caption{
    \textbf{Qualitative comparison on additional scenes of the DTU benchmark~\cite{jensen2014large}.}
    \name can produce both smooth surfaces and detailed structures compared to prior work,
    despite limited improvement in simply textured and highly reflective objects.
    }
    \label{fig:app_dtu_additional_results}
\end{figure*}

%% file: figures/appendix_rendering_dtu.tex
\begin{figure*}[tb]
    \centering
    \includegraphics[width=\linewidth]{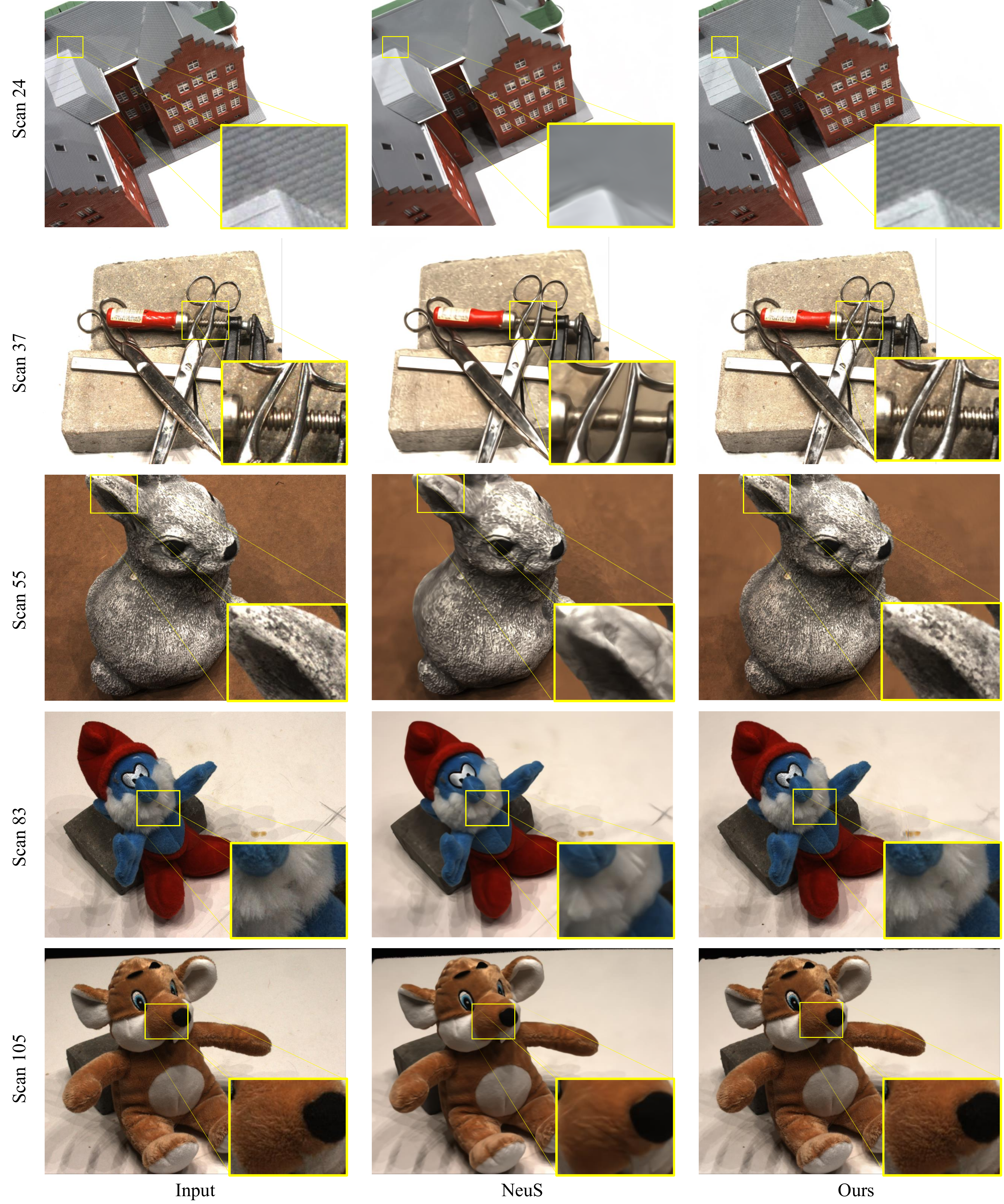}
    \caption{
    \textbf{Qualitative comparison of RGB image synthesis on the DTU benchmark~\cite{jensen2014large}. }
    Compared to NeuS~\cite{wang2021neus}, \name generates high-fidelity renderings with minute details.
    }
    \label{fig:app_dtu_rendering}
\end{figure*}

%% file: tables/dtu_mask.tex
\setlength\tabcolsep{2.6em}
\begin{table}[b]
\centering
\ra{1.1}
\resizebox{\linewidth}{!}{%
\begin{tabular}{@{}lcc}
\specialrule{.2em}{.1em}{.1em}
 & \multicolumn{2}{c@{}}{Chamfer distance (mm) $\downarrow$} \\ \cline{2-3}
 & IDR masks & Our masks \\ \specialrule{.13em}{.1em}{.1em}
NeuS~\cite{wang2021neus} & 1.48 & 0.99 \\
NeuralWarp~\cite{darmon2022improving} & 1.20 & 0.73  \\
Ours & 1.29 & 0.76 \\
\specialrule{.2em}{.1em}{.1em}
\end{tabular}%
}
\caption{
\textbf{Quantitative results on Scan 83 of the DTU dataset~\cite{jensen2014large}} using object masks provided by IDR~\cite{yariv2020multiview} and annotated by us. }
\label{tab:dtu_mask}
\end{table}

%% file: figures/appendix_dtu_mask.tex
\begin{figure*}[tb]
    \centering
    \includegraphics[width=\linewidth]{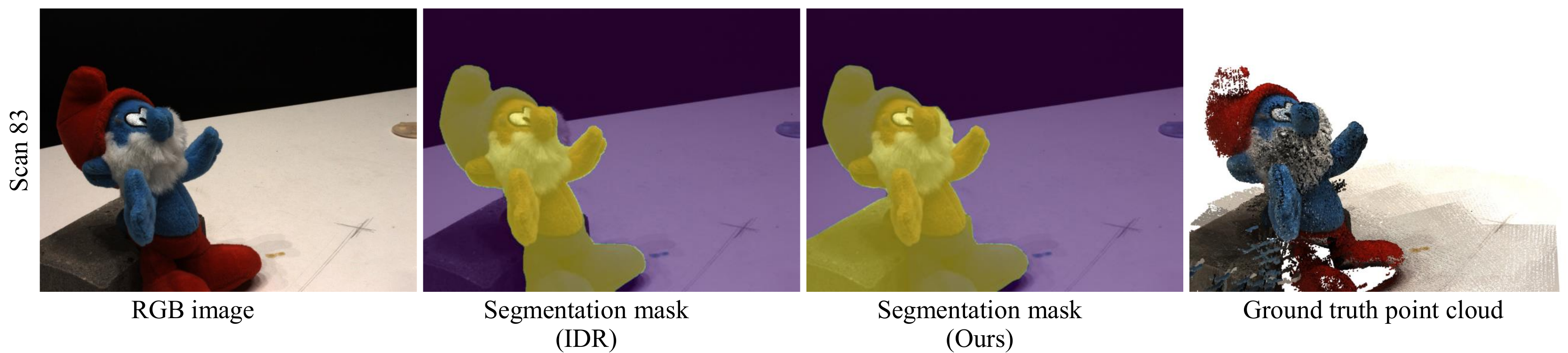}
    \caption{
    \textbf{We manually re-annotate the foreground object masks of the DTU dataset.}
    We note that the object masks provided by IDR miss the objects partially on Scan 83. 
    The IDR masks also do not include the bricks holding objects, while ground truth point clouds have the brick.
    Our updated segmentation masks fix the above issues for better evaluation.
    }
    \label{fig:dtu_mask}
\end{figure*}

%% file: figures/appendix_per_img_embed.tex
\begin{figure}[tb]
    \centering
    \includegraphics[width=\linewidth]{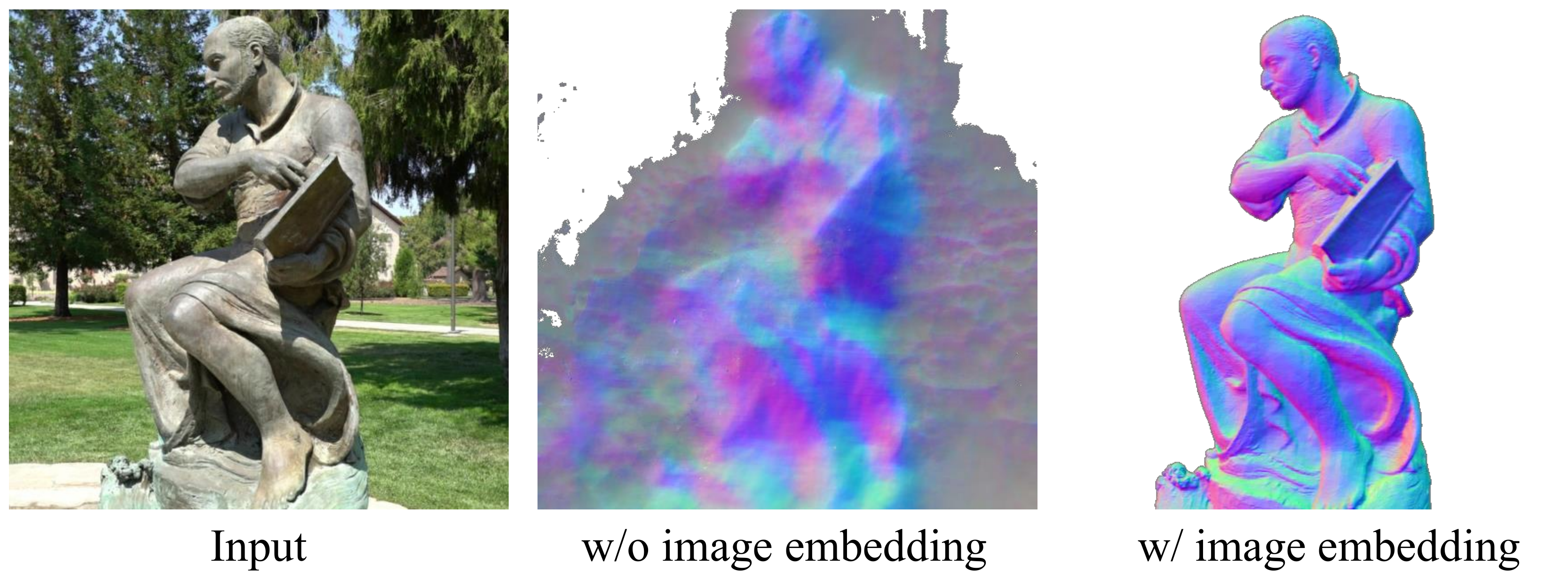}
    \caption{
    \textbf{Qualitative comparison of normal maps without and with per-image embedding.}
    Floaters are greatly reduced with per-image embedding.
    }
    \label{fig:app_per_img_embed}
\end{figure}

%% file: figures/appendix_curv_decay.tex
\begin{figure}[tb]
    \centering
    \includegraphics[width=\linewidth]{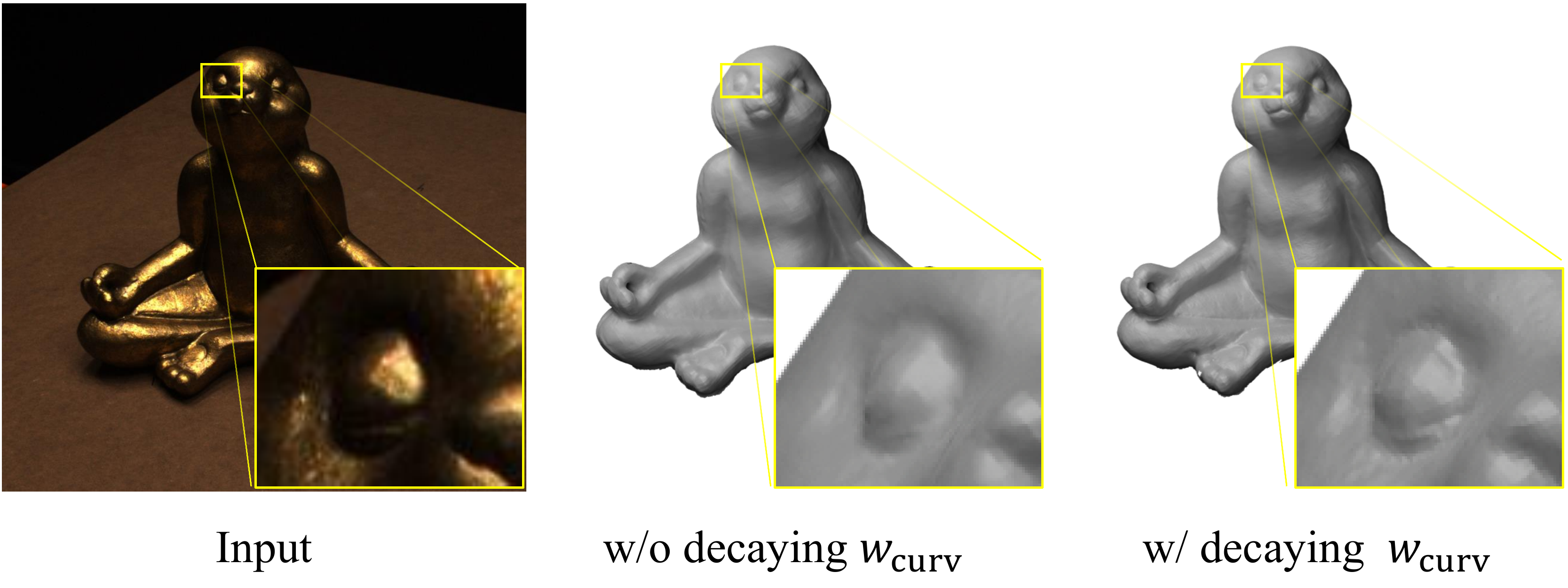}
    \caption{
    \textbf{Qualitative comparison of without and with decaying $w_\text{curv}$.}
    Decaying $w_\text{curv}$ reduces the regularization strength as $\epsilon$ decreases, thus preserving details better.
    }
    \label{fig:app_curv_decay}
\end{figure}

%% file: figures/appendix_normal_comparison.tex
\begin{figure}[tb]
    \centering
    \includegraphics[width=\linewidth]{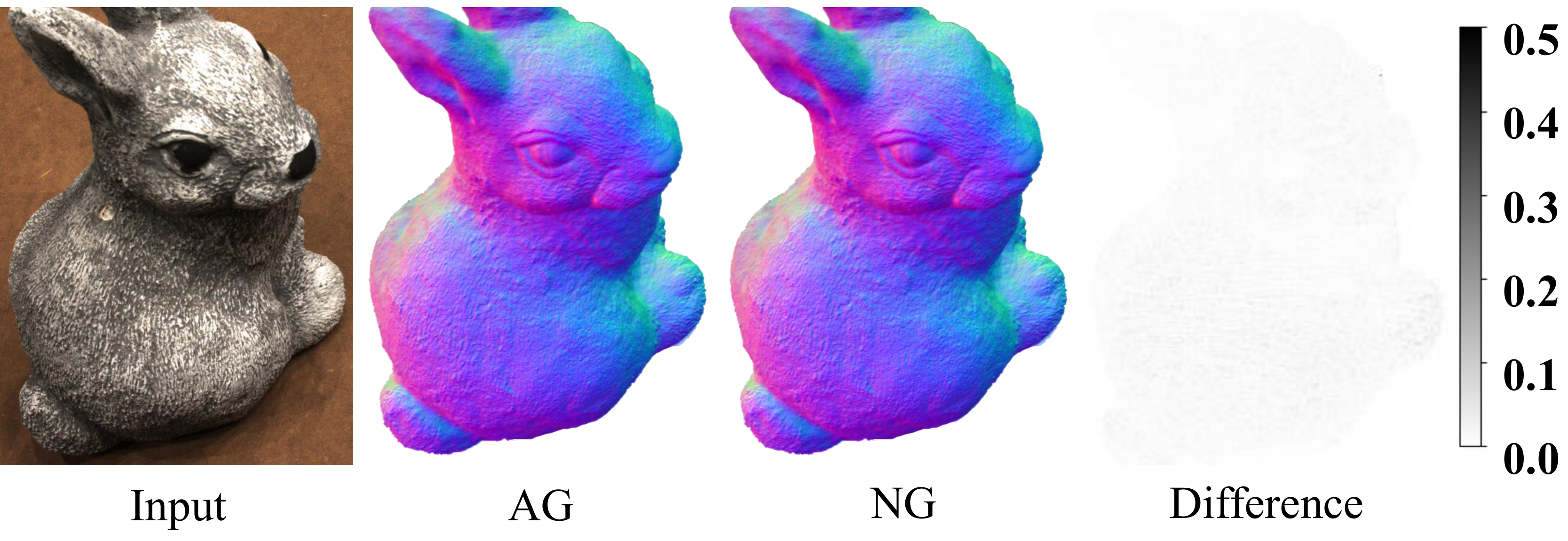}
    \caption{
    \textbf{Qualitative visualizations of surface normals computed from analytical gradient (\ag) and numerical gradient (\fd).}
    The results are nearly identical at the end of the optimization due to the small step size $\epsilon$.
    }
    \label{fig:app_normal_comparison}
\end{figure}

%% file: figures/appendix_color_network.tex
\begin{figure*}[tb]
    \centering
    \includegraphics[width=\linewidth]{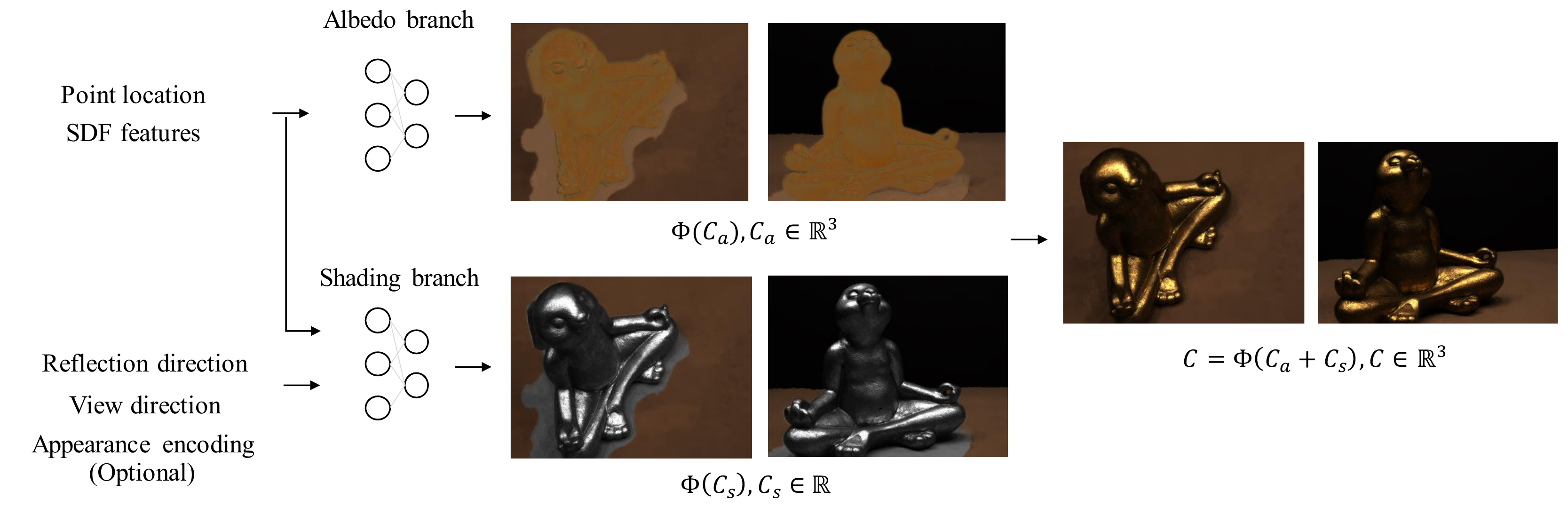}
    \caption{Color network design for intrinsic decomposition. The decomposition scheme includes albedo and shading images.}
    \label{fig:intrinsic_decomp}
\end{figure*}

%% file: figures/appendix_color_result.tex
\begin{figure*}[tb]
    \centering
    \includegraphics[width=\linewidth]{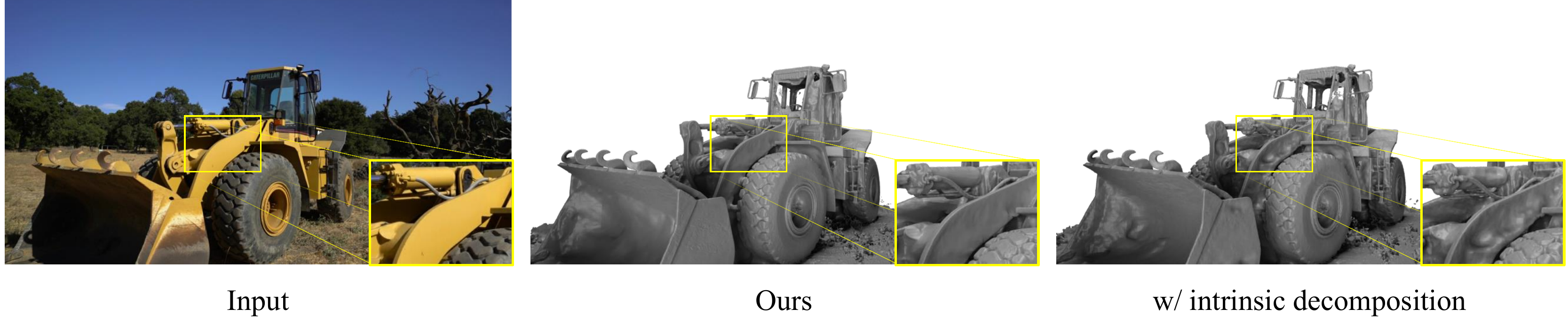}
    \caption{
    Qualitative comparison of different color network designs.
    We find that the intrinsic decomposition we implemented lacks smoothness in regions with homogeneous color, while the color network proposed by IDR~\cite{yariv2020multiview} produces smooth surfaces.
    }
    \label{fig:color_result}
\end{figure*}

%% file: tables/speed.tex
\setlength\tabcolsep{1.7em}
\begin{table}[tb]
\centering
\resizebox{\linewidth}{!}{%
    \begin{tabular}{@{}lcc@{}}
    \specialrule{.2em}{.1em}{.1em}
     & Training time (s) & Inference time (s) \\
     \specialrule{.1em}{.1em}{.1em}
    NeuS~\cite{wang2021neus} & 0.16 & 0.19\\
    \fd (Ours) & 0.12 & 0.08\\
    \ag & 0.10 & 0.08 \\
    \specialrule{.2em}{.1em}{.1em}
    \end{tabular}
}
\caption{
\textbf{Computational time comparison between NeuS~\cite{wang2021neus}, \ag and \fd using Nvidia V100 GPUs.}
Training time reported is per iteration and inference time reported is for surface extraction of $128^3$ resolution.
There is approximately a 1.2 times slowdown in training time of ours compared to \ag.
Ours is still faster than NeuS due to the smaller-sized MLP used.
For inference time, both ours and AG are more than 2 times faster than NeuS.}
\label{tab:appendix_speed}
\end{table}